\documentclass[11pt]{article}

\usepackage[final]{acl}

\usepackage{times}
\usepackage{latexsym}

\usepackage[T1]{fontenc}

\usepackage[utf8]{inputenc}

\usepackage{microtype}

\usepackage{inconsolata}

\usepackage{graphicx}

\usepackage{hyperref}
\usepackage{url}
\usepackage{graphicx}
\usepackage{caption}
\usepackage{enumitem}
\usepackage{wrapfig}
\usepackage{array} 
\usepackage{booktabs}
\usepackage{multirow}
\usepackage{amsmath}
\usepackage{cleveref}
\usepackage{colortbl}
\usepackage{xcolor}
\usepackage{subcaption}

\usepackage{xspace}
\usepackage{pifont}
\usepackage{multirow}
\usepackage{multicol}
\usepackage{tabularx}
\usepackage[most]{tcolorbox}
\usepackage[dvipsnames]{xcolor}
\usepackage{tabularx}
\usepackage{verbatim}
\usepackage{fancyvrb}
\usepackage{booktabs} 
\tcbuselibrary{listings, breakable}
\newcommand{\cmark}{\textcolor{OliveGreen}{\ding{51}}}
\newcommand{\xmark}{\textcolor{BrickRed}{\ding{55}}}

\crefname{section}{Sec.}{Sections}
\crefname{figure}{Fig.}{Figures}
\crefname{table}{Tab.}{Tables}
\crefname{equation}{Equation}{Equations}
\crefname{appendix}{Appendix}{Appendixs}

\title{GOAT: A Training Framework for Goal-Oriented Agent with Tools}

\author{
Hyunji Min$^{1}$ \quad
Sangwon Jung$^{2}$ \quad
Junyoung Sung$^{1}$ \quad \\
\textbf{Dosung Lee$^{1}$ \quad
Leekyung Han$^{1}$ \quad
Paul Hongsuck Seo$^{1}$} \\
$^{1}$Korea University \quad
$^{2}$Trillion Labs\\
\texttt{\{daream2, jys7451, dslee1219, happilee, phseo\}@korea.ac.kr} \\
\texttt{sangwon.jung@trillionlabs.co}
}

\begin{document}
\maketitle
\begin{abstract}
Large language models (LLMs) have evolved from pure text generators into interactive agents capable of invoking external tools.
However, LLM agents still struggle with \textit{goal-oriented} queries, which require decomposing high-level objectives into sequences of interdependent API calls with accurate planning and execution.
Current approaches rely on zero-shot evaluation due to the absence of training data; while proprietary models such as GPT-4 exhibit strong reasoning capabilities, smaller open-source models remain ineffective at complex tool use.
To address this limitation, we propose a novel training framework \textbf{GOAT}, that enables fine-tuning LLM agents without human annotation. 
GOAT automatically synthesizes goal-oriented API execution data from API documents using a novel \textbf{call-first} generation paradigm, that constructs training data based on executed API call sequences.
Through extensive experiments, we show that GOAT-trained agents achieve state-of-the-art performance across multiple existing goal-oriented benchmarks. In addition, we introduce \textbf{GOATBench}, a new goal-oriented API execution benchmark, and demonstrate that agents trained with GOAT also excel in this setting.
These results highlight GOAT as a practical path toward building robust open-source LLM agents capable of complex reasoning and tool use. Our dataset and code are available at: \url{https://github.com/KU-MIIL/GOAT}.
\end{abstract}

\begin{figure*}[t]
\centering
\includegraphics[width=1\textwidth]{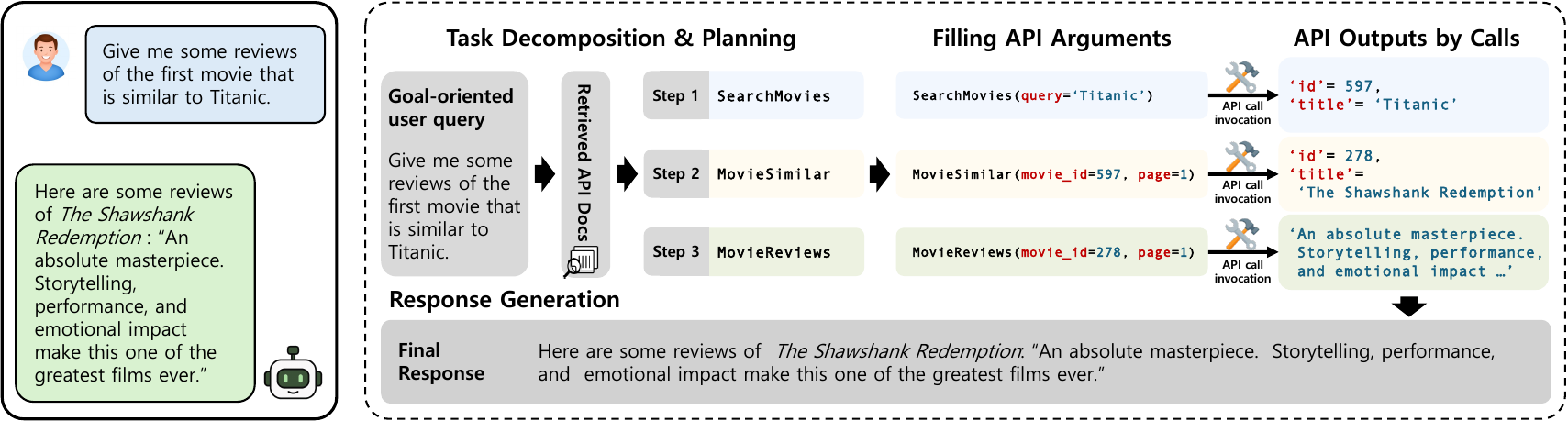}
\caption{\textbf{Goal-oriented API execution task.}
To solve a goal-oriented user query, the LLM agent performs step-by-step task planning, executes a sequence of interdependent API calls, and generates a natural language response. The figure illustrates the workflow where the user query is decomposed into subtasks, mapped to API calls, and each function call is executed by filling API arguments based on the outputs of previous responses, ultimately yielding a coherent answer.
}
\label{fig:main_task}
\vspace{-1.5em}
\end{figure*}
\section{Introduction}

Recent advances in large language models (LLMs) have led to remarkable progress across a wide range of natural language processing tasks~\citep{zhao2025surveylargelanguagemodels,achiam2023gpt}.
Beyond pure language understanding, a growing line of research explores a new paradigm, where LLMs act as \textit{agents} that can actively interact with the external world~\citep{kim24openvla,intelligence2025pi_,qin2025ui,han2025dialnav} including tool(API) call to respond to user needs~\citep{qin2024toollearningfoundationmodels,Patil2023GorillaLL,Schick2023ToolformerLM}.
This paradigm, referred to as \textit{tool learning}, moves LLMs beyond text generation toward task planning, tool invocation, and answer generation.



While prior work on tool learning mainly considers simpler settings, such as single-step queries~\citep{Patil2023GorillaLL, Schick2023ToolformerLM} or queries with explicitly specified instructions~\citep{Qin2023ToolLLMFL, liu2024agentbench, Shen2023TaskBenchBL}, we focus on more realistic and challenging scenarios, which we term \textbf{goal-oriented tasks}.
As illustrated in~\cref{fig:main_task}, a goal-oriented query specifies only a high-level objective rather than detailed step-by-step instructions, requiring the agent to break down the objective into intermediate steps, determine which APIs to call, and infer appropriate arguments for the selected API functions. Consequently, these tasks demand strong reasoning for task decomposition, long-horizon planning, and call realization that captures the interdependency between APIs, followed by their coordinated execution.



However, progress on goal-oriented tasks has been limited by the lack of training data: constructing datasets that capture inter-API dependencies needs human annotation and is prohibitively costly at scale.
Consequently, current approaches~\citep{Song2023RestGPTCL} mainly rely on zero-shot evaluation, expecting models to perform complex reasoning without task-specific supervision.
To address this challenge, we propose a novel training framework for \textbf{G}oal-\textbf{O}riented \textbf{A}gent with \textbf{T}ools \textbf{(GOAT)}, which enables fine-tuning LLM agents in the absence of human-annotated data through a fully automatic synthetic data generation pipeline.
Our approach builds on the common practice that once a target set of APIs is specified, their function documentation is already available and can be directly leveraged to construct training data.
Specifically, given API documents of a target environment, GOAT first induces an API dependency graph through a refinement pipeline, retaining only feasible invocation relations.
GOAT then samples connected subgraphs representing interdependent subtasks and follows a \textbf{call-first} generation: it instantiates and executes the corresponding APIs before generating a goal-oriented user query and final response aligned with the executed API sequence.
Finally, GOAT jointly fine-tunes the LLM and retriever on the resulting dataset, enabling them to reason over interdependent APIs and produce coherent responses.

Furthermore, leveraging our data generation pipeline together with human labeling, we curate \textbf{GOATBench}, an evaluation benchmark for goal-oriented tasks.
Through extensive experiments on goal-oriented benchmarks including GOATBench, we show that GOAT enables open-source models to achieve state-of-the-art performance in multiple benchmarks, in some cases even surpassing certain closed-source models with strong reasoning capabilities. Our main contributions are as follows:

\begin{itemize}[itemsep=2pt, topsep=0pt, parsep=0pt, partopsep=0pt, leftmargin=*]
\item We propose \textbf{GOAT}, a novel training framework that automatically constructs goal-oriented API execution datasets from the target API documents without human annotation, enabling efficient domain-specific adaptation of LLM agents and strengthening their reasoning  capabilities.
\item Through extensive experiments on goal-oriented benchmarks, we demonstrate that agents trained with GOAT achieve state-of-the-art performance with open-source models.
\item We introduce \textbf{GOATBench}, a new large-scale evaluation benchmark for goal-oriented tasks, and confirm consistent performance gains of GOAT-trained agents on this benchmark as well.
\end{itemize}

\section{Related Work}

\noindent \textbf{Task Formulations in Tool Learning} \ \ 
Tool learning tasks range from simple scenarios, where a few APIs are directly provided in the prompt~\citep{yao2023react,Schick2023ToolformerLM,ViperGPT,zhuang2023toolqa}, to more advanced formulations involving many available tools, where models must retrieve the relevant one.
Among these, some benchmarks provide queries with explicit step-by-step instructions, leaving little need for planning or multi-step reasoning~\citep{Patil2023GorillaLL,Qin2023ToolLLMFL,HuggingGPT,yang2023gpt4tools,Shen2023TaskBenchBL,liu2024agentbench}. 
In contrast, the most challenging and realistic setting is the \textit{goal-oriented task}, where users provide only high-level goals and requires planning and executing sequences of interdependent API calls.
Prior work, such as RestBench and API-Bank~\citep{Song2023RestGPTCL,li2023apibank}, relies heavily on costly human annotation to identify API dependencies and construct call paths, which limits their scalability and restricts their use primarily to evaluation.
In this work, we propose \textbf{GOAT}, a fully automatic framework that constructs goal-oriented data without human annotation.
Building on GOAT, we introduce \textbf{GOATBench}, a large-scale benchmark that requires only lightweight human verification, thereby alleviating the scalability and coverage limitations of prior manually annotated benchmarks.

\begin{table*}[t]
\centering
\scalebox{0.75}{
\begin{tabular}{lcccccc}
\toprule
Work & Real API & Fully Automatic & Scalable & API Call Dependency & Goal-oriented Query \\
\midrule
ToolFormer~\citep{Schick2023ToolformerLM} & \xmark & \cmark & \xmark & \xmark & \xmark \\
Gorilla~\citep{Patil2023GorillaLL} & \cmark & \cmark & \cmark & \xmark & \xmark \\
ToolLLM~\citep{Qin2023ToolLLMFL} & \cmark & \cmark & \cmark & \xmark & \xmark \\
API-Bank*~\citep{li2023apibank} & \xmark & \cmark & \xmark & \xmark & \xmark \\
TaskBench~\citep{Shen2023TaskBenchBL} & \cmark & \xmark & \xmark & \cmark & \xmark \\
ToolFlow~\citep{Wang2024ToolFlowBL} & \cmark & \cmark & \cmark & \cmark & \xmark \\
Magnet~\citep{magnet} & \cmark & \cmark & \cmark & \cmark & \xmark \\
ToolDial~\citep{shim2025tooldialmultiturndialoguegeneration} & \cmark & \cmark & \cmark & \cmark & \xmark \\
\midrule
\textbf{Ours (GOAT)} & \cmark & \cmark & \cmark & \cmark & \cmark \\
\bottomrule
\end{tabular}
}
\caption{\textbf{Comparison of existing synthetic training data generation works in tool learning.} 
Our work uniquely generates synthetic training data specifically targeting goal-oriented tasks. *Note that although API-Bank provides a benchmark that includes goal-oriented tasks, its training data generation process does not target such queries.}
\label{tab:related_work}
\vspace{-0.5em}
\end{table*}


\noindent \textbf{Synthetic Training Data for Tool Learning} \ \
High-quality training data is essential for developing LLM agents that can reliably use tools in real-world scenarios.
Early efforts generated single-API training data by pairing a function with a corresponding query–api call pair~\citep{Schick2023ToolformerLM,Patil2023GorillaLL}, and later extended to multi-API settings by random sampling multiple APIs and generating a combined parallel query~\citep{Qin2023ToolLLMFL,li2023apibank}.
To better handle multiple tool use, graph-based approaches construct API dependency graphs and sample connected sequences, generate fine-grained instructions describing multi-step API calls~\citep{Shen2023TaskBenchBL} or multi-turn dialogues~\citep{Wang2024ToolFlowBL,magnet,shim2025tooldialmultiturndialoguegeneration}.
However, the dependencies modeled in these graph-based frameworks are relatively weak, as they are derived from simple heuristics or algorithms without validating whether the connections are truly meaningful, and none of the above generate holistic goal-oriented queries that require agents to plan over multiple interdependent API calls. 
In contrast, our approach automatically constructs such goal-oriented data, thereby addressing this gap and enabling effective training for realistic tool-use scenarios (see~\cref{tab:related_work}). 
Moreover, many pipelines follow an \textit{instruction-first} strategy~\citep{wang-etal-2023-self-instruct,Patil2023GorillaLL,Qin2023ToolLLMFL,tang2023toolalpaca,li2023apibank},  where an LLM first generates a user query from API functions, and then produces API calls with arguments corresponding to that query. 
This leads to a self-reinforcing bias, producing data dominated by ``easy'' cases already solvable by LLM and offering limited benefit for learning complex reasoning.
In contrast, we adopt a \textit{call-first} strategy where we instantiate executable API workflows directly from documentation then abstract the workflow into a natural-language query.

\section{A training framework for \textbf{G}oal-\textbf{O}riented \textbf{A}gent with \textbf{T}ools}

We propose a novel human-annotation-free training framework for Goal-Oriented Agents with Tools (GOAT), motivated by the prohibitive cost of collecting manual annotations. 
We begin with the practical observation that agents are typically deployed in specific domains with a fixed set of target APIs. 
This makes it reasonable to assume that API documentation for these APIs is available beforehand.
Leveraging this assumption, our framework (i) automatically generates training samples from the available API documentation, and (ii) fine-tunes an LLM and retriever model on these samples to strengthen its goal-oriented reasoning capabilities for the target API environment.

\begin{figure*}[t]
\centering
\includegraphics[width=1\textwidth]{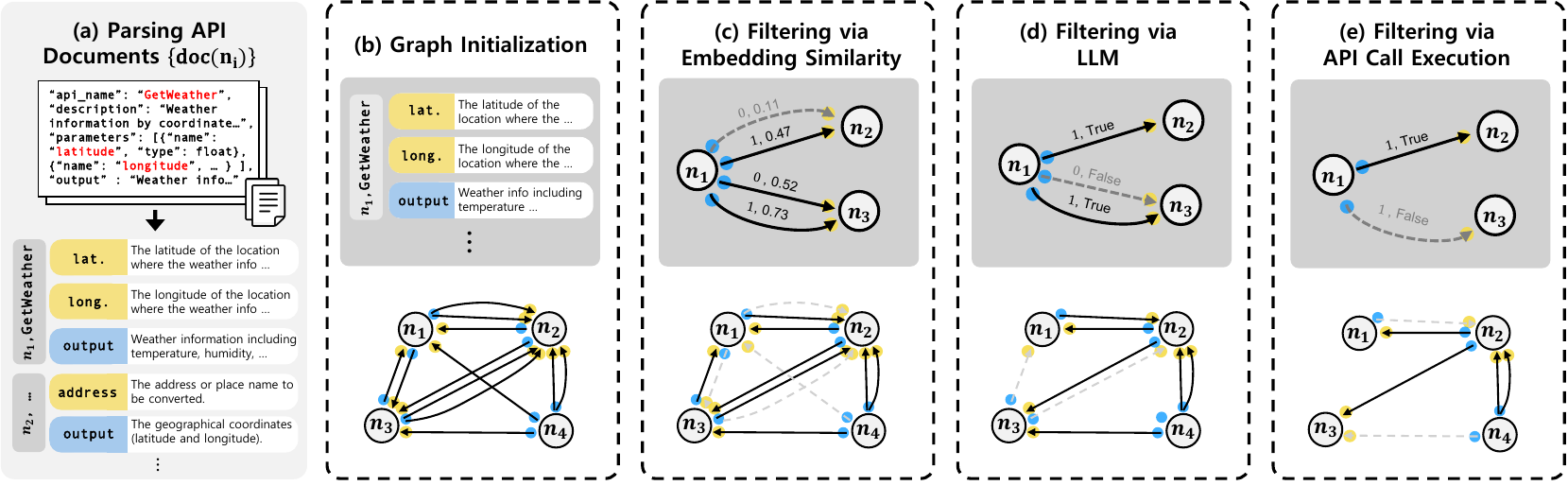}
\vspace{-2em}
\caption{\textbf{The overview of API dependency graph construction process.}
Given the API documents, each document is first parsed to extract function descriptions, which are then used to initialize a raw dependency graph in (a). 
This graph is progressively refined through three filtering steps (c)-(e), resulting in the final API dependency graph that captures reliable relations among APIs.
The graphs shown under (b)-(e) illustrate how the API dependency graph evolves as it is progressively refined through each filtering step.
}
\label{fig:graph_construction}
\vspace{-1em}
\end{figure*}

\subsection{Automatic Dataset Construction}
\label{subsec:auto_data_construction}
A central challenge in training goal-oriented agents is the absence of manually annotated training datasets or synthetic training datasets with task-specific supervision.
Our framework addresses this challenge by fully automatically constructing synthetic training data, eliminating the need for costly human annotations.
It is important to note that in goal-oriented tasks, intermediate API calls are inherently interdependent, as earlier calls are often executed to prepare inputs for later ones. The generated synthetic data must therefore capture this interdependency across API calls. Such dependency information is typically implicit in API documentation, and we leverage this dependency cues during the data generation process.
Specifically, we construct goal-oriented API execution data through two main stages. First, given a set of API documents describing API functions, we build an API dependency graph that captures all possible ways in which the output of one API can serve as an input to another (\cref{subsec:graph_construction}). 
From the resulting graph, we extract connected subgraphs to create synthetic data points, each consisting of a goal-oriented user query $u$, a set of call units $\{(s, c, o)\}$—where $s$ is a sub-query, $c$ the corresponding API call, and $o$ its output—and the final response $r$ (\cref{subsec:data_construction}).

\subsubsection{API Dependency Graph Construction}
\label{subsec:graph_construction}
Given a set of API documents $\{\text{doc}(n_i)\}$ with description of API functoin $n_i$ and its implicit dependency information,  we formulate an API dependency graph $G=(\mathcal{V}, \mathcal{E})$, where $\mathcal{V}=\{n_i\}$ represents the set of API functions and an edge in $\mathcal{E} = \{(n_i, n_j, k)\}$ indicates that the output of an API function $n_i$ can be used as the $k$-th argument in a subsequent call to $n_j$. Since a single output may serve multiple parameters for the same API function, $G$ is a multidigraph that allows multiple directed edges between the same pair of nodes.
The dependency graph captures the input–output dynamics across APIs, reflecting the execution flow for goal-oriented tasks.


To construct a dependency graph $G$, we leverage the strong reasoning capabilities of LLMs. A high-performing LLM generates candidate arguments for a source API, executes them, and verifies whether the resulting outputs can populate the parameters of a destination API. Exhaustively applying this procedure to all input–output pairs, however, is prohibitively expensive.
To balance reliability and efficiency, we instead start from an over-complete graph and progressively reduce LLM usage via a staged filtering pipeline. Inexpensive coarse filters first remove clearly incompatible edges, while only a shrinking subset is escalated to increasingly precise and costly checks. The effect of each stage is detailed in Appendix~\ref{appendix_filter}. The overall process is illustrated in \cref{fig:graph_construction}.

\begin{figure*}[t]
\centering
\includegraphics[width=1\textwidth]{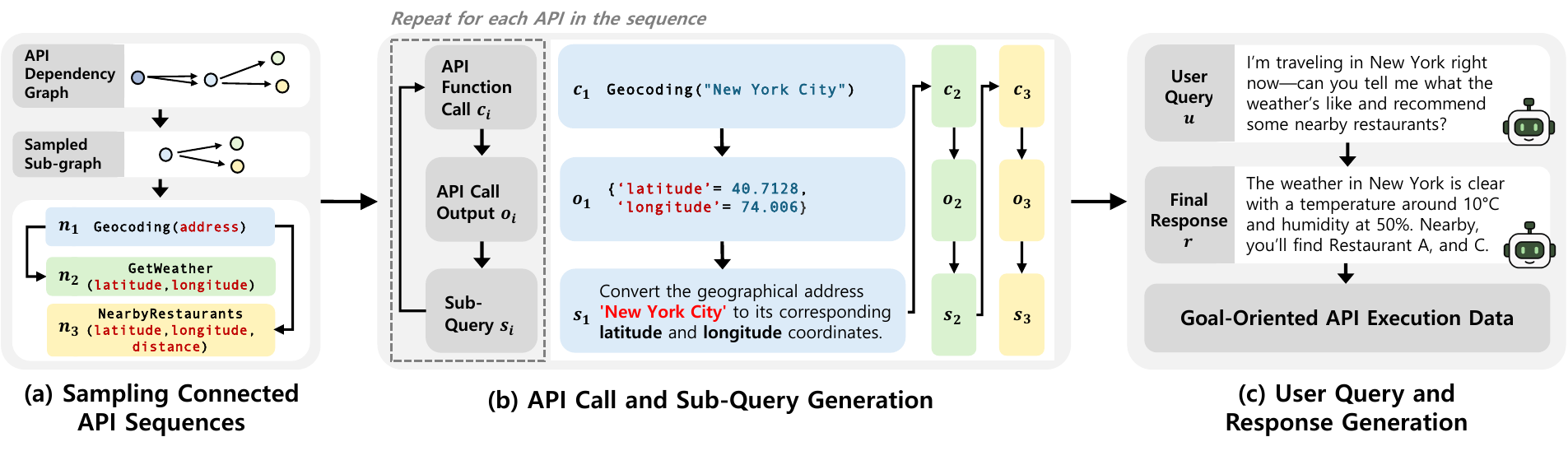}
\caption{\textbf{Overview of goal-oriented API execution data construction.} 
The process involves (a) sampling connected API sequences, (b) generating API calls, outputs, and sub-queries, and (c) composing user queries and final responses.}
\label{fig:data_construction}
\vspace{-1.5em}
\end{figure*}

\noindent\textbf{API Document Parsing and Graph Initialization} \ \ 
We begin by extracting the input and output specifications of each API function $n_i$ by parsing its document $\text{doc}(n_i)$, which define the endpoints for edges in the graph (see \cref{fig:graph_construction}a). 
Each specification includes the language description of input parameters and output; we denote the description of the $k$-th input of $n_i$ as $\text{In}(n_i, k)$ and its output description as $\text{Out}(n_i)$.
These specifications are extracted using an LLM with the prompt shown in \cref{fig:doc_parsing}, and the resulting structured representations are illustrated at the bottom of \cref{fig:graph_construction}a and detailed further in Appendix~\ref{appendix_parsing}.
Using these extracted specifications, we initialize an fully connected multidigraph by adding directed edges from the output of each API function to every input parameter of every other API function (see \cref{fig:graph_construction}b). This over-complete graph serves as the starting point for subsequent filtering and refinement.


\noindent\textbf{Filtering via Embedding Similarity} \ \ 
To prune unlikely edges, we compare the source API output descriptions with the destination API input descriptions extracted in the previous step (\cref{fig:graph_construction}c). For each edge $(n_i, n_j, k)$, we compute the cosine similarity between Sentence-BERT (SBERT) \citep{reimers-2019-sentence-bert} embeddings of $\text{Out}(n_i)$ and $\text{In}(n_j, k)$, and discard edges with similarity below a threshold $\tau$. Because this step does not require LLM invocations, it provides an efficient means of eliminating clearly incompatible pairs.
To avoid over-pruning, we set $\tau$ conservatively, favoring recall over precision, since more precise filtering is applied in later stages.

\noindent\textbf{Filtering via LLM} \ \ 
As illustrated in \cref{fig:graph_construction}d, edges that survive the similarity filter are further evaluated using a single LLM call per edge. 
Given API documents ($\text{doc}(n_i)$ and $\text{doc}(n_j)$) and the descriptions ($\text{In}(n_j, k)$ and $\text{Out}(n_i)$), the LLM determines whether the output of $n_i$ can meaningfully populate the $k$-th input of $n_j$ using the prompt in Appendix~\ref{llm filtering prompt}. For edges deemed valid, the LLM also generates a natural-language justification explaining the semantic compatibility between the output and the input parameter. 
We reuse these justifications later in data construction (\cref{subsec:data_construction}) to guide argument generation.
This step assesses \textit{semantic plausibility} based solely on descriptions, without grounding in actual values.

\noindent\textbf{Filtering via API Call Execution} \ \ 
As in Figure~\ref{fig:graph_construction}e, candidate edges are finally validated through execution with three LLM calls per edge. Unlike prior semantic-only check, this step grounds edges in \textit{concrete values} obtained from real API executions. 
For each edge $(n_i, n_j, k)$, the LLM first generates plausible arguments to execute a source API call $c_i$ for $n_i$ and obtain its output $o_i$. 
It then constructs a destination API's call $c_j$ for $n_j$, filling the $k$-th argument with content extracted from $o_i$, and finally verifies whether $c_j$ is coherent and executable as a continuation of $c_i$ (Appendix~\ref{api call filtering prompt}). 
By validating edges through actual executions, this step removes semantically plausible but non-executable connections, ensuring that the final dependency graph represents reliable API workflows.


\subsubsection{Goal-Oriented API Execution Data Construction}
\label{subsec:data_construction}

Once the dependency graph is finalized, we construct goal-oriented task samples through a three-step process.
First, since the dependency graph is composed of all possible API relationships, we extract connected subgraphs and treat them as candidate API sequences. Second, for each sequence, we sequentially instantiate and execute each API functions by filling in their input arguments, executing them in order, and passing API outputs forward as inputs to subsequent calls. In doing so, we also generate sub-queries that explain each API call in natural language. Finally, based on the sub-queries and the outputs obtained from calling a API sequence, we construct a user query that captures the overall intent, and then generate a natural final response that interprets the outputs in their full context to address this query. The result is a dataset of samples where every component---from query to intermediate reasoning to final response---is grounded in real API executions.
The full process is illustrated in \cref{fig:data_construction}.

\noindent\textbf{Sampling Connected API Sequences} \ \ 
As illustrated in \cref{fig:data_construction}a, we begin by extracting all possible \textit{connected} subgraphs from $G$, with up to $L=4$ nodes, each of which serves as the basis of a task instance. If the subgraph contains cycles, we randomly break edges, producing an acyclic structure. The resulting acyclic multidigraph is then topologically sorted to produce an execution order $(n_{k_1}, n_{k_2},\dots,n_{k_L})$ where $k_l$ represents the $l$-th node index among the $L$ nodes. This guarantees that APIs are invoked in a dependency-respecting sequence, forming a valid workflow.

\noindent\textbf{API Call and Sub-Query Generation} \ \ 
Following the predetermined sequence, each $n_{k_\ell}$ is instantiated as an API call $c_{k_\ell}$ by filling the input arguments (Appendix~\ref{api call sequence generation prompt}). 
For each argument, the LLM either (i) extracts a value from the output $o_{k_m}$ of a preceding API call $(m < \ell)$ when a dependency edge exists, or (ii) synthesizes a plausible value from the function specification $\text{doc}(n_{k_\ell})$ otherwise.
Since outputs are often complex and nested, non-trivial reasoning over the execution context is required; here, we prompt the LLM with the justification produced during the second filtering stage (\cref{subsec:graph_construction}) to guide correct value selection. 
Once all arguments are filled, the API call is executed to obtain $o_{k_\ell}$. 
Alongside each API call, we generate a natural-language sub-query $s_{k_\ell}$ that reflects the operation performed by $c_{k_\ell}$.
This sub-query is constructed from the API specification, instantiated arguments, and execution context, ensuring alignment with both the invoked function and its role in the overall task flow (Appendix~\ref{subinstruction generation prompt}). Applying this procedure iteratively over the ordered sequence yields a complete execution trajectory $\{(s, c, o)\}$ grounded in real API execution (\cref{fig:data_construction}b).

\noindent\textbf{User Query and Response Generation} \ \ 
\cref{fig:data_construction}c depicts the final stage, where once the full set of call units $\{(s, c, o)\}$ has been constructed, we generate a goal-oriented user query $u$ that encapsulates the overall task. The query is created by summarizing the sub-queries $\{s\}$ and abstracting the high-level objective that the entire API workflow is designed to achieve, as detailed in Appendix~\ref{query generation}. Finally, we generate the final response $r$ corresponding to the user query $u$. For this, we provide the LLM with $u$ together with the full set of triplets ${(s, c, o)}$, as described in Appendix~\ref{final response generation}. This allows the model to produce a coherent answer grounded in the outputs of the composed API calls.

\begin{table*}[t]
\centering
\scalebox{0.75}{
\begin{tabular}{lllccccccc}
    \toprule
    &  & Prompting & GOAT
    & \multicolumn{3}{c}{TMDB} 
    & \multicolumn{3}{c}{Spotify} \\
    \cmidrule(lr){5-7} \cmidrule(lr){8-10}
    & Backbone & Method &  FT
    & Success\% $\uparrow$ & CP\% $\uparrow$ & $\Delta$ Len. $\downarrow$
    & Success\% $\uparrow$ & CP\% $\uparrow$ & $\Delta$ Len. $\downarrow$ \\
    \specialrule{1pt}{2pt}{2pt}

    \multirow{3}{*}{\textcolor{gray}{Closed-source}} 
        & \textcolor{gray}{text-davinci-003} & \textcolor{gray}{Baseline}   & - & \textcolor{gray}{29.0} & \textcolor{gray}{33.0} & \textcolor{gray}{+1.52} & \textcolor{gray}{14.5} & \textcolor{gray}{36.4} & \textcolor{gray}{+1.10} \\
    & \textcolor{gray}{text-davinci-003} & \textcolor{gray}{ReAct}       & - & \textcolor{gray}{44.0} & \textcolor{gray}{57.0} & \textcolor{gray}{+0.76} & \textcolor{gray}{54.5} & \textcolor{gray}{49.1} & \textcolor{gray}{+0.31} \\
    & \textcolor{gray}{text-davinci-003} & \textcolor{gray}{RestGPT}     & - & \textcolor{gray}{75.0} & \textcolor{gray}{79.0} & \textcolor{gray}{+0.55} & \textcolor{gray}{72.7} & \textcolor{gray}{74.5} & \textcolor{gray}{+0.25} \\
    \midrule
    \multirow{6}{*}{Open-source} 
    & Llama2-13B & RestGPT    & - & 0.0 & 0.0  & -    & 0.0 & 0.0 & - \\
    & Llama2-13B & Baseline   & \xmark & 0.0 & 0.0  & -    & 3.5 & 7.0 & +0.00 \\
    & Llama2-13B  & Baseline  & \cmark & \textbf{7.0} & \textbf{13.0} & +0.71 & \textbf{28.1} & \textbf{28.1} & +0.44 \\
    \cmidrule(lr){2-10}
    & Vicuna-13B & RestGPT    & - & 9.0 & \textbf{15.0} & +1.21 & 12.7 & 20.6 & +1.52 \\
    & Vicuna-13B & RestGPT*    & - & 1.0 & 0.0  & +0.00  & 0.0 & 0.0 & - \\
    & Vicuna-13B & Baseline   & \xmark & 0.0 & 0.0  & -     & 0.0 & 0.0 & - \\
    & Vicuna-13B & Baseline  & \cmark & \textbf{17.0} & 14.0 & +0.53 & \textbf{29.8} & \textbf{33.3} & +1.00 \\
    \bottomrule
\end{tabular}
}
\caption{\textbf{Experiment results on RestBench.} Closed-source and RestGPT results are reported numbers from original paper, shown here for reference. Metrics are Success\%, Correct Path\%, and $\Delta$ Solution Length. For Vicuna-13B, we additionally reproduced RestGPT using the released code and found substantially lower performance than reported (marked with * in the table).}
\label{tab:main_restbench}
\end{table*}


\begin{table*}[t]
\centering
\scalebox{0.75}{
\begin{tabular}{lllcccccc}
    \toprule
     & Prompting Method & Backbone & FT Method & Success\% $\uparrow$ & CP\% $\uparrow$ & Correctness\% $\uparrow$ & ROUGE & \\
    \specialrule{1pt}{2pt}{2pt}
    \multirow{3}{*}{\textcolor{gray}{Closed-source}} 
        & \textcolor{gray}{API-Bank} & \textcolor{gray}{GPT-3 Davinci} & \textcolor{gray}{-} & \textcolor{gray}{-}  & \textcolor{gray}{-}  & \textcolor{gray}{0.00} & \textcolor{gray}{0.0156} \\
        & \textcolor{gray}{API-Bank} & \textcolor{gray}{GPT-3.5-turbo} & \textcolor{gray}{-} & \textcolor{gray}{-} & \textcolor{gray}{-} & \textcolor{gray}{22.00} & \textcolor{gray}{0.3809} \\
        & \textcolor{gray}{API-Bank} & \textcolor{gray}{GPT-4} & \textcolor{gray}{-} & \textcolor{gray}{-} & \textcolor{gray}{-} & \textcolor{gray}{70.00} & \textcolor{gray}{0.4808} \\
    \midrule
    \multirow{5}{*}{Open-source}  
        & API-Bank   & Alpaca-7B   & - & -  & -  & 0.00 & 0.0860 \\
        & API-Bank  & ChatGLM-6B  & - & -  & -  & 0.00 & 0.1522 \\
        \cmidrule(lr){2-8}
        & API-Bank & Llama-7B & API-Bank & - & - & 20.00 & 0.3425 \\
        & Baseline & Llama-7B & - & 0.0 & 0.0 & 0.00 & 0.0048 \\
        & Baseline & Llama-7B & \textbf{GOAT} & \textbf{38.0} & \textbf{42.0} & \textbf{42.22} & 0.3173 \\
    \bottomrule
\end{tabular}
}
\caption{\textbf{Experiment results on API-Bank.} Performance of the API-Bank prompting method is reported numbers from original paper. Since the official inference code is unavailable, additional metrics (Success\%, Correct Path\%) could not be evaluated.}
\label{tab:main_apibank}
\vspace{-1em}
\end{table*}


    

\begin{table*}[t]
\centering
\scalebox{0.75}{
\begin{tabular}{lllccccccc}
    \toprule
     &  &  &  
     & \multicolumn{3}{c}{Inter Tool} 
     & \multicolumn{3}{c}{Single Tool}\\
     
     \cmidrule(lr){5-7} \cmidrule(lr){8-10}
     & Prompting Method & Backbone & FT Method & SA & IA & SR & SA & IA & SR \\
    \specialrule{1pt}{2pt}{2pt}
    \multirow{1}{*}{\textcolor{gray}{Closed-source}} 
        & \textcolor{gray}{Baseline} & \textcolor{gray}{GPT-4.1} & \textcolor{gray}{-}
        & \textcolor{gray}{22.7} & \textcolor{gray}{10.9} & \textcolor{gray}{27.4} 
        & \textcolor{gray}{45.4} & \textcolor{gray}{21.5} & \textcolor{gray}{51.3} 
     \\
    \midrule
    \multirow{5}{*}{Open-source} 
        & ReACT & Llama2-7B & - & 1.5 & 0.2 & 1.0 & 2.2 & 0.0 & 1.9 \\
        & ReACT & Llama2-7B & ToolLLM & 13.1 & 0.9 & 1.0 & 28.7 & 0.5 & 3.9 \\
        & ReACT & Llama2-7B & \textbf{GOAT} & \textbf{33.2} & \textbf{4.8} & \textbf{1.9} & \textbf{41.0} & \textbf{7.4} & \textbf{7.2} \\
        \cmidrule(lr){2-10}
        & Baseline & Llama3-8B & - & 9.7 & 2.9 & 7.2 & 18.2 & 6.6  & 7.9  \\
        & Baseline & Llama3-8B & \textbf{GOAT} & \textbf{59.4} & \textbf{26.5} & \textbf{12.3} & \textbf{69.1} & \textbf{33.8} & \textbf{16.5} \\
        
    \bottomrule
\end{tabular}
}
\caption{\textbf{Experiment results on GOATBench.} Metrics are SA (Selection Accuracy), IA (Invocation Accuracy), and SR (Success Rate).}
\label{tab:main_goatbench}
\vspace{-1em}
\end{table*}

\begin{table}[t]
\centering
\scalebox{0.75}{
\begin{tabular}{lcc}
    \toprule
    Data Generation Method & TMDB & Spotify \\
    \specialrule{1pt}{2pt}{2pt}
    Instruction-first & 5.0 & 17.5 \\
    Call-first & \textbf{7.0} & \textbf{28.1} \\
    \bottomrule
\end{tabular}
}
\caption{\textbf{Comparison of data generation strategies on RestBench.}
Success Rate\% under different data generation methods, evaluated using a baseline prompting approach with a Llama-2-13B backbone.}
\label{tab:main_callfirst}
\vspace{-1.5em}
\end{table}


\subsubsection{Discussions}
\label{discussion}

A fundamental limitation of existing data synthesis approaches is that they rely on models to infer API calls from natural language query during data generation.
Since it is the precise capability we aim to improve in LLMs, \textbf{instruction-first} generation suffers from self-reinforcing bias: models tend to produce correct calls only for queries they already handle well, yielding data skewed toward ``easy'' instances and limiting its effectiveness for learning complex reasoning. Consequently, instruction-first generation has primarily been used for distillation, transferring knowledge from larger closed models to smaller open models.
To address this issue, we adopt a \textbf{call-first strategy}.
Instead of inferring API calls from language, we first instantiate concrete API calls from API documentation, execute them to obtain outputs $(c, o)$, and then generate a natural language query that summarizes the completed workflow.
This direction of generation is substantially easier for LLMs, as it requires abstraction over given inputs and outputs rather than complex planning and inference.
By leveraging this asymmetry, call-first generation converts the summarization and abstraction strengths of LLMs into reliable supervision for the inverse task, enabling effective learning of mapping high-level user goals to executable, interdependent API calls.
We empirically demonstrate the advantages of call-first generation over instruction-first alternatives in ~\cref{results}.

\subsection{Agent Training}
On top of synthetic goal-oriented dataset, we train LLM agent consisting of a language model and an API retriever. 
The language model is instruction-tuned with supervision from the generated goal-oriented API execution data. 
Specifically, the model learns to plan the correct sequence of calls, fill in their arguments, and generate the final natural-language response by aggregating the execution outputs. 
To improve generalization beyond specific argument patterns, argument values are masked during training.  
The retriever is trained on ground-truth query–API document pairs, enabling it to map user queries to the relevant API specifications.  
Together, this framework enables effective learning of LLM agents for domain-specific API tasks, where both planning and execution are grounded in real API behavior.

\vspace{-0.2cm}
\section{Experiments}

\subsection{Experiment Setup}
\label{training_details}
We evaluate LLM Agents on goal-oriented API execution benchmarks where no human-annotated training data is available. 
For each benchmarks, we use Llama-3-70B-Instruct~\citep{llama3-70b-instruct} to construct the corresponding synthetic training data within GOAT.
While training, the LLM is fine-tuned with Low-Rank Adaptation (LoRA) method~\citep{lora} to enable parameter-efficient fine-tuning on API execution tasks.
For the retriever, we fine-tune a dense retrieval model based on the SBERT architecture~\citep{reimers-2019-sentence-bert}, specifically using the all-MiniLM-L6-v2 encoder using InfoNCE loss.
All hyperparameters and implementation details are provided in Appendix~\ref{appendix_implementation}.

As a \textit{Baseline} for prompting agents, we followed the decomposition-first method from prior work~\citep{huang2024understandingplanningllmagents}; 
where given a user query, the agent retrieves the top-$k=5$ relevant API documents and predicts the entire sequence of API calls in a single step. 
The planned sequence is executed iteratively, with each call incorporating outputs of previous ones, and the final answer is composed from the collected results. 
This baseline was the strongest among those we tested and was thus chosen as our main baseline, with additional results for the others provided in Appendix~\ref{appendix_llmagent}.



\subsection{Evaluation Benchmarks}

\noindent \textbf{RestBench}~\citep{Song2023RestGPTCL} \ \ 
RestBench is a human-generated benchmark with two test sets built on real-world APIs from TMDB (movie database) and Spotify (music streaming). The evaluation is based on three metrics: (i) \textit{Success\%}, assessed by human evaluation to determine whether the output fulfills the user request; 
(ii) \textit{Correct Path\%}, which counts a case as correct if the gold API call path appears as a subsequence of the model-generated path, considering only the API function sequence and ignoring parameter values; 
and (iii) $\Delta$ \textit{Solution Length}, defined as the mean number of additional API calls, measured over successful cases only, reflecting plan efficiency.

\noindent \textbf{API-Bank}~\citep{li2023apibank} \ \ 
API-Bank consists of human-implemented Python APIs with manually annotated queries and API call paths.  It includes three task sets, among which we focus on the \textit{Plan+Retrieve+Call} set, as it uniquely reflects the goal-oriented query setting involving multi-API and multi-call reasoning. The official evaluation metrics include (i) \textit{Correctness}, measured by the precision of API call responses, and (ii) \textit{ROUGE}, computed between the model’s final response and the gold response. Since these metrics do not directly capture overall task success, we additionally report \textit{Success\%} and \textit{Correct Path\%}, following the definitions in RestBench.

\noindent \textbf{GOATBench} \ \ 
GOATBench is a human-verified benchmark on real APIs from RapidAPI, constructed through GOAT pipeline with additional human curation (see~\cref{goatbench} for details). 
Tasks are categorized into \textit{Single Tool} and \textit{Inter Tool}, depending on whether multiple APIs come from the same tool or from different tools.
For evaluation, we adopt three commonly used metrics:  
(i) \textit{API Selection Accuracy (SA)}, measuring the Jaccard similarity between the predicted and ground-truth sets of API functions~\citep{Wang2024mtubench};  
(ii) \textit{API Invocation Accuracy (IA)}, similar to SA but requiring all arguments to match in addition to the function name; and  
(iii) \textit{Success Rate (SR)}, a GPT-based evaluation metric that determines whether the final answer sufficiently and correctly solves the user query given the tool execution results~\citep{Qin2023ToolLLMFL}.  
For SR evaluation, we employ GPT-4.1 and provide the exact evaluation prompt in~\cref{fig:Success Rate Prompt}. 



\vspace{-0.2cm}
\subsection{Results}
\label{results}

%
\noindent \textbf{Results on RestBench} \ \ 
\cref{tab:main_restbench} shows experimental results on RestBench. Without fine-tuning, baseline results with open-source models are nearly zero in most cases, indicating complete failure.
This underscores the necessity of fine-tuning when building on-premise agents with open-source models.
GOAT training yields clear and consistent improvements over these baselines with both Llama2-13B and Vicuna-13B achieving substantial gains after fine-tuning.
The fine-tuned models not only surpass the previously reported open-source state-of-the-art method, RestGPT, but in some cases even outperform the closed-source model.

%


\noindent \textbf{Results on API-Bank} \ \ 
\cref{tab:main_apibank} further demonstrates the effectiveness of GOAT through consistent gains on API-Bank. 
Fine-tuning with GOAT improves performance over zero-shot Llama-7B and, in some cases, even surpasses closed-source models. 
Compared to the original API-Bank training setup~\citep{li2023apibank}, which relies on synthetic non-executable API functions and instruction-first data generation, GOAT achieves stronger results by aligning supervision directly with the target task.
While most evaluation metrics improve, ROUGE is slightly lower; however, as shown in \cref{fig:qual_ex_2}, it is less informative as LLMs may hallucinate fluent responses without correct API execution.





\noindent \textbf{Results on GOATBench} \ \ 
As shown in \cref{tab:main_goatbench}, GOAT training achieves substantially higher performance compared to models without training.
We evaluate its effectiveness with an additional prompting method, ReACT~\citep{yao2023react}, and find that GOAT consistently improves performance under both ReACT and the baseline prompting setup.
Notably, the gains hold not only over the non-fine-tuned model but also over the fine-tuned one on the instruction-first data generated by ToolLLM~\citep{Qin2023ToolLLMFL}, operating with ReACT.
Although ToolLLM also generates synthetic data using the same APIs as GOATBench, its data construction process is not tailored to goal-oriented queries but instead emphasizes parallel multi-API queries with fine-grained instructions, resulting in poor performance on the goal-oriented evaluation.

\noindent \textbf{Ablation on Call-First Data Generation} \ \ 
To empirically demonstrate the advantage of the call-first data generation strategy, we compare it against an instruction-first baseline on RestBench.
Both methods are evaluated using a baseline prompting approach with a Llama-2-13B backbone.
As shown in Table~\ref{tab:main_callfirst}, model trained with call-first generated data consistently achieves higher success rates than instruction-first generation.
These results indicate that constructing supervision from executable API calls provides more reliable training signals for learning goal-oriented API reasoning.

\section{Conclusions}
In this work, we present \textbf{GOAT}, a fully automatic training framework for equipping LLM agents with goal-oriented reasoning capabilities over interdependent APIs.
Unlike prior approaches that rely solely on zero-shot evaluation in the absence of training data, GOAT exploits existing API documentation to automatically construct synthetic supervision.
This allows open-source LLMs to be fine-tuned efficiently while retaining strong generalization across diverse goal-oriented scenarios.
Furthermore, leveraging GOAT pipeline, we curated \textbf{GOATBench}, a benchmark for evaluating goal-oriented tool use.
Extensive experiments on GOATBench and other benchmarks demonstrate that GOAT-trained agents not only achieve state-of-the-art performance among open-source models but also, in some cases, surpass closed-source systems with strong reasoning ability.
Overall, our results highlight GOAT as a practical and scalable path toward building robust open-source LLM agents that can reason over complex tool interactions and deliver accurate responses to high-level user goals.


\section*{Limitations}
While our framework yields noticeable performance improvements in goal-oriented API execution, it has several limitations. Although we observe gains even on unseen APIs, GOAT is primarily designed for in-domain agent training, and its generalization to unseen scenarios remains limited. Additionally, while our approach is robust to typical documentation variability, the quality of API documents can influence the accuracy of the generated data, as our method reflects the information available in the source descriptions.

\section*{Ethics Statement}
We acknowledge the importance of transparency and responsible use of publicly available resources in our research. All datasets, benchmarks, models, and APIs employed in this study were publicly released for academic research and were used in full compliance with their respective licenses and intended purposes. 
Specifically, we used the Meta-Llama-3-8B-Instruct model under the Llama 3 Community License, the Sentence-BERT model under the Apache 2.0 License, and publicly available benchmarks such as RestGPT and StableToolBench, as well as the TMDB API under its non-commercial usage terms. 
No proprietary or private data were used and this research does not involve human subjects or sensitive personal information.
To promote transparency and responsible research, we will publicly release the full GOAT suite, including the GOATBench dataset, the data generation pipeline, and the training and evaluation code.
\section*{Acknowledgements}
This work was the result of project supported by KT(Korea Telecom)-Korea University AICT R\&D Center.
This work was supported by the IITP grants (RS-2025-02653113, IITP-2026-RS-2020-II201819, IITP-2026-RS2024-00436857, RS-2024-00398115, IITP-2026-RS-2025-02304828, RS-2026-25507282), the NRF grant (RS-2025-23523979) and the KOCCA grant (RS-2024-00345025) funded by the Korean government (MSIT, MSCT).

\bibliography{custom}

\clearpage
\appendix
\begin{figure*}[t]
\centering
\includegraphics[width=1\textwidth]{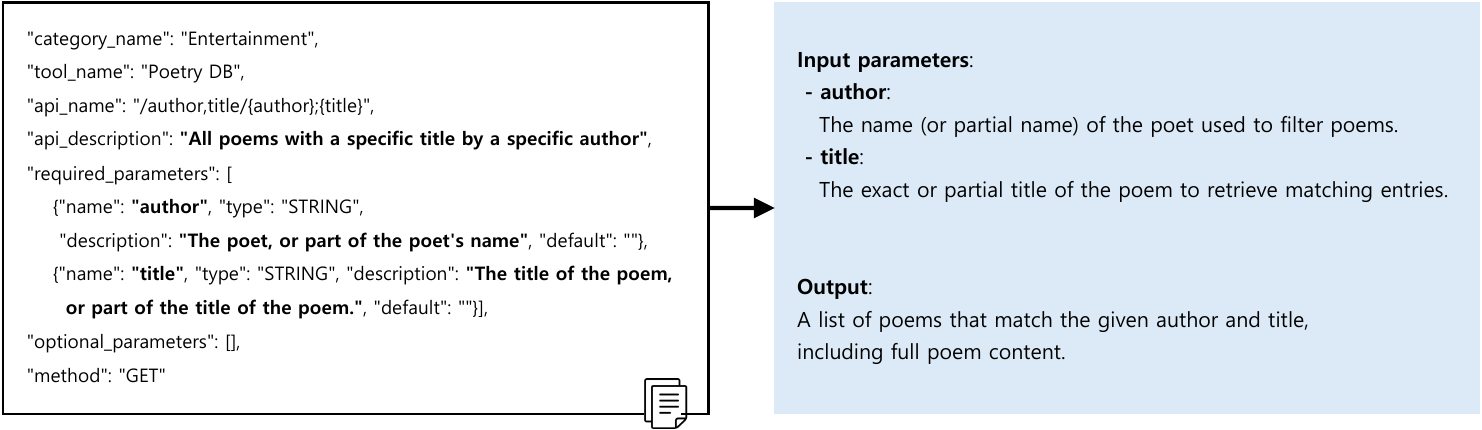}
\caption{\textbf{Example of API document parsing result.}}
\label{fig:doc_parsing}
\end{figure*}
\section{API Dependency Graph Construction}
\subsection{API Document Parsing Example}
\label{appendix_parsing}
\cref{fig:doc_parsing} illustrates an example of how raw API document is converted into structured representations.
By prompting LLM as in \cref{fig:doc_parsing}, we extract both the types and number of input parameters of each API function together with their semantic roles, as well as the semantic meaning of the returned output.

\begin{table*}[t]
    \centering
    \begin{tabular}{l c c c}
        \hline
         & Embedding Similarity & \  \  LLM \   \ & API Call Execution \\
        \hline
        Precision & 0.25 & 0.59 & 0.90 \\
        Recall & 0.92 & 0.42 & 0.36 \\
        \hline
    \end{tabular}
    \caption{\textbf{Precision and recall of each pruning stage.}}
    \label{tab:pruning_results}
\end{table*}

\subsection{Filtering Valid Dependency Edges}
\label{appendix_filter}

GOAT adapt three-step filtering process to obtain valid API dependency edges.
The precision and recall of each pruning stage are summarized in~\cref{tab:pruning_results}. 
The first row represents the precision, while the second row corresponds to the recall, based on 500 data edges that we manually annotated.
The results demonstrate that our method successfully identifies valid edges $e = (n_i, n_j, k)$ while filtering out spurious connections, thereby enhancing the quality of the constructed graph.  

\subsection{API Dependency Graph Example}
\begin{figure*}[t]
\centering
\includegraphics[width=1\textwidth]{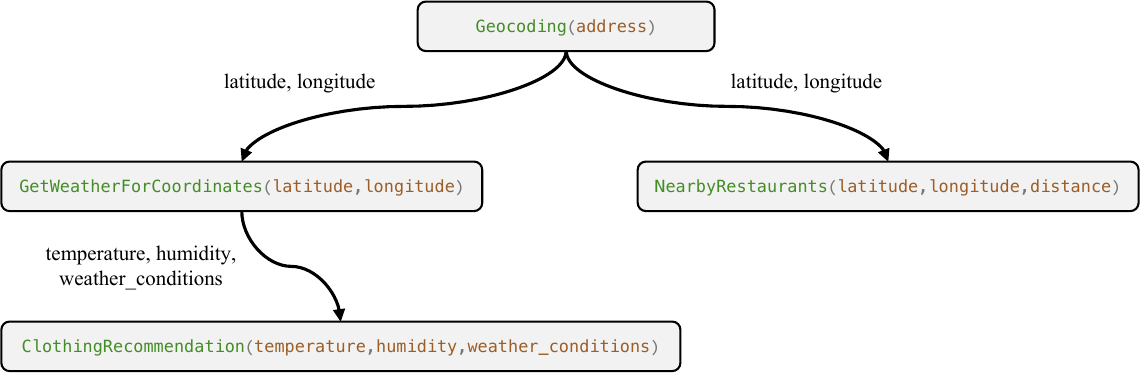}
\caption{\textbf{Example of Constructed API Dependency Graph from APIBank APIs.} 
}
\label{fig:graphexample}
\end{figure*}

\cref{fig:graphexample} shows example of constructed API dependency graph.

\section{Implementation Details}
\label{appendix_implementation}
We use Llama-3-70B-Instruct model to generate synthetic training data for each benchmark within GOAT.
While filtering via embedding similarity, we set the hyperparameter $\tau$ to 0.2 for GOATBench and APIBank, 
while a more conservative threshold of 0.05 is adopted for RestBench.
Fine-tuning of LLM model was performed on a single NVIDIA H100 GPU for 3 epochs on every experiment.
We adopted the Low-Rank Adaptation (LoRA) method \cite{lora} with \( r = 8 \), $\alpha=16$, and a dropout ratio of 0.05 to enable parameter-efficient fine-tuning on instruction-guided API execution tasks. 
All experiment results reported in this paper are based on a single run without variation across random seeds.

We additionally provide the data statistics used in each training stage. API calls occur during dependency-graph construction (roughly two calls per edge) and during synthetic data generation (one call per node), and all endpoints used in our experiments (TMDB, Spotify, and StableToolBench APIs) are free services. The number of synthetic instances used for LLM fine-tuning is 8570 for RestGPT–TMDB, 924 for RestGPT–Spotify, 108 for API-Bank, and 1631, 1354, 650, and 420 for the Entertainment, Financial, Food, and Travel subsets of GOATBench. The corresponding numbers of query–document pairs used for SBERT training are 33169 for RestGPT–TMDB, 3389 for RestGPT–Spotify, 180 for API-Bank, and 5091, 4752, 1957, and 1166 for the four GOATBench domains.

\section{GOAT-Generated Data Quality Analysis}
As we discussed in ~\cref{discussion}, GOAT enables benchmark curation with much less human effort than other fully human annotated benchmarks such as RestBench~\cite{Song2023RestGPTCL} and API-Bank~\cite{li2023apibank}.
\cref{fig:qual_data,fig:qual_data2,fig:qual_data3} show qualitative comparsion between GOAT-generated data and fully human-generated data examples. 

\label{quality_analysis}
\begin{figure*}[t]
\begin{tcolorbox}[colframe=gray!50!black, colback=gray!10!white, title=GOAT-generated Example, width=\textwidth, sharp corners=southwest]
\scriptsize
\begin{verbatim}
{
    "query": "Get the keywords for movie 'The Shawshank Redemption' released in 1994.",
    "api_path": [
        {
            "api_name": "GET /search/movie",
            "input": {
                "query": "The Shawshank Redemption",
                "include_adult": false,
                "region": "US",
                "year": 1994,
                ...
            },
            "output": {
                "error": "",
                "response": {
                    "page": 1,
                    "results": [
                        {
                            ...
                            "id": 278,
                            "original_title": "The Shawshank Redemption",
                            ...
                        }
                    ],
                    ...
                }
            },
            "sub_instruction": "Search for movies with the query 
            'The Shawshank Redemption' in the US region, released in 1994, 
            on the first page, excluding adult content."
        },
        {
            "api_name": "GET /movie/{movie_id}/keywords",
            "input": {
                "movie_id": 278
            },
            "output": {
                "error": "",
                "response": {
                    "id": 278,
                    "keywords": [
                        {"id": 378, "name": "prison"},
                        {"id": 6054, "name": "friendship"},
                        {"id": 796, "name": "police brutality"},
                        {"id": 417, "name": "corruption"},
                        {"id": 818, "name": "based on novel or book"},
                        {"id": 6089, "name": "freedom"},
                        ...
                    ]
                }
            },
            "sub_instruction": "Retrieve the keywords associated with the movie 
            'The Shawshank Redemption' which has an ID of 278."
        }
    ],
    "final_response": "The keywords for the movie 'The Shawshank Redemption' released in 
    1994 are: prison, friendship, police brutality, corruption, based on novel or book, 
    freedom, and more."
}
\end{verbatim}
\end{tcolorbox}
\begin{tcolorbox}[colframe=gray!50!black, colback=gray!10!white, title=Human-generated Example, width=\textwidth, sharp corners=southwest]
\scriptsize
\begin{verbatim}
{
      "query": "What are some common keywords associated with movie \"Titanic\"?",
      "solution": [
          "GET /search/movie",
          "GET /movie/{movie_id}/keywords"
      ]
}
\end{verbatim}
\end{tcolorbox}
\caption{\textbf{Comparison of Goat-Generated and Human-generated on TMDB.}}
\label{fig:qual_data}
\end{figure*}
\begin{figure*}[t]
\begin{tcolorbox}[colframe=gray!50!black, colback=gray!10!white, title=GOAT-generated Example, width=\textwidth, sharp corners=southwest]
\scriptsize
\begin{verbatim}
{
    "query": "Get me an album of the first artist I'm following.",
    "api_path": [
        {
          "api_name": "GET /me/following",
          "input": {
            "type": "artist",
            "after": "",
            "limit": 20
          },
          "output": [
            {
                ...
                "total": 3,
                "items": [
                  {
                    "id": "74ASZWbe4lXaubB36ztrGX",
                    "name": "Bob Dylan",
                    ...
          ]
            "sub_instruction": "Retrieve the current user's followed artists, with a 
            maximum of 20 results, using the 'artist' ID type and starting from the 
            beginning."
        },
        {
          "api_name": "GET /artists/{id}/albums",
          "input": {
            "id": "74ASZWbe4lXaubB36ztrGX",
            ...
          },
          "output": [
            {
              ...
              "items": [
                {
                  "id": "6y2WHyqRUCeHrjMXvjnRmD",
                  "name": "Modern Times",
                  "album_type": "album",
                  "total_tracks": 20,
                  "release_date": "2006-08-29",
                  ...,
                  "artists": [
                    {
                      "id": "74ASZWbe4lXaubB36ztrGX",
                      "name": "Bob Dylan",
                      "external_urls": {
                        "spotify": "https://open.spotify.com/artist/74ASZWbe4lXaubB36..."
                      }
                ...
          ]
            "sub_instruction": "Retrieve the album information for the artist with id
            '74ASZWbe4lXaubB36ztrGX', which is the first artist in the list of followed 
            artists, with a limit of 20 items and an offset of 0, without any specific 
            market or album type filter."
        }
    ],
    "final_response": "The first artist you are following is Bob Dylan. One of his albums 
    is 'Modern Times' (released on August 29, 2006), which contains 20 tracks. You can 
    explore more of his albums on Spotify."
}
\end{verbatim}
\end{tcolorbox}
\begin{tcolorbox}[colframe=gray!50!black, colback=gray!10!white, title=Human-generated Example, width=\textwidth, sharp corners=southwest]
\scriptsize
\begin{verbatim}
{
    "query": "Recommend more artists base on my first following artist",
    "solution": [
        "GET /me/following",
        "GET /artists/{id}/related-artists"
    ]
}
\end{verbatim}
\end{tcolorbox}
\caption{\textbf{Comparison of Goat-generated and Human-Generated on Spotify.}}
\label{fig:qual_data2}
\end{figure*}
\begin{figure*}[t]
\begin{tcolorbox}[colframe=gray!50!black, colback=gray!10!white, title=GOAT-generated Example, width=\textwidth, sharp corners=southwest]
\scriptsize
\begin{Verbatim}
{
    "query": "What are the tax deductions for a Data Scientist's salary?",
    "api_path": [
        {
            "api_name": "GetOccupationSalary",
            "input": {
                "occupation": "Data Scientist"
            },
            "output": "{\"salary\": 150000}",
            "sub_instruction": "Retrieve the salary for the occupation 'Data Scientist'."
        },
        {
            "api_name": "TaxCalculator",
            "input": {
                "salary": 150000.0
            },
            "output": "{\"salary_after_tax\": 105000.0}",
            "sub_instruction": "Calculate the tax deductions for a salary of 150000.0."
        }
    ],
    "final_response": "As a Data Scientist, your take-home salary after tax deductions 
    would be around $105,000, based on a gross salary of $150,000."
}
\end{Verbatim}
\end{tcolorbox}
\begin{tcolorbox}[colframe=gray!50!black, colback=gray!10!white, title=Human-generated Example, width=\textwidth, sharp corners=southwest]
\scriptsize
\begin{Verbatim}
{
    "query": "Calculate the monthly salary after taxes for a Data Scientist",
    "api_path": [
        {
            "api_name": "GetOccupationSalary",
            "input": {
                "occupation": "Data Scientist"
            },
            "output": {
                "api_name": "GetOccupationSalary",
                "input": {
                    "occupation": "Data Scientist"
                },
                "output": {
                    "salary": 150000
                },
                "exception": null
            }
        }
        {
            "api_name": "TaxCalculator",
            "input": {
                "salary": "150000"
            },
            "output": {
                "api_name": "TaxCalculator",
                "input": {
                    "salary": 150000.0
                },
                "output": {
                    "salary_after_tax": 105000.0
                },
                "exception": null
            }
        }
    ],
    "final_response": "For a Data Scientist with a salary of $150,000, the monthly salary 
    after taxes is $105,000.",
}
\end{Verbatim}
\end{tcolorbox}
\caption{\textbf{Comparison of Goat-generated and Human-Generated on APIBank.}}
\label{fig:qual_data3}
\end{figure*}

\begin{table*}[t]
\centering
\begin{tabular}{lccc}
\toprule
 & \textbf{GOATBench} & \textbf{RestBench} & \textbf{API-Bank} \\
\midrule
\# of Domains   & 4   & 2   & 8   \\
\# of APIs      & 830 & 94  & 73(21)  \\
\# of Instances & 660 & 157 & 314(50) \\
Avg. Len        & 3.1 & 2.41 & 2.91 \\
\bottomrule
\end{tabular}
\caption{\textbf{Statistical comparison of GOATBench, RestBench, and API-Bank.} Numbers in parentheses denote statistics for the Plan+Retrieve+Call subset of API-Bank. }
\label{tab:benchmark_stats}
\end{table*}

\begin{table}[t]
\centering
\begin{tabular}{cc}
\toprule
\textbf{Subgraph Size} & \textbf{\# of Instances} \\
\midrule
2 & 93  \\
3 & 417 \\
4 & 150 \\
\bottomrule
\end{tabular}
\caption{\textbf{Distribution of subgraph sizes in GOATBench.}}
\label{tab:subgraph_sizes}
\end{table}

\begin{figure*}[t]
\begin{tcolorbox}[colframe=gray!50!black, colback=gray!10!white, title=GOATBench Data Example, width=\textwidth, sharp corners=southwest]
\scriptsize
\begin{Verbatim}
{
    "query": "Can you help me find a hotel near the Flixbus stations in Paris?",
    "api_path": [
        {
            "api_name": "Flixbus_v2/Autocomplete",
            "input": {
                "query":"Paris",
                "locale":"en"
            },
            "output": {[
              {
                "id": "dcc05c30-9603-11e6-9066-549f350fcb0c",
                "name": "Paris CDG Airport",
                "address": "Gare Routière Roissypole",
                ...
              },
              {
                "id": "7f6214cd-95d7-4bbf-8283-55af226fa33a",
                "name": "Paris (Pont de Sèvres)",
                ...
              },
              ...
            ]},
            "subinstruction": "Search for Flixbus stations using the query 'Paris'
            and retrieve results in English."
        },
        {
            "api_name": "Hotels/locations/v2/search",
            "intput": {
                "query":"Paris CDG Airport",
                "locale":"en_US",
                "currency":"EUR"
            },
            "output": {[
              {
                "name": "Paris Marriott Charles de Gaulle Airport Hotel",
                "geoId": "5854",
                ...
              },
              {
                "name": "Holiday Inn Express Paris - CDG Airport, an IHG Hotel",
                "geoId": "19497017",
                ...
              },
              ...
            ]},
            "subinstruction": "Find hotels near Paris CDG Airport station."
        }
    ],
    "final_response": "There is a Flixbus stop at Paris CDG Airport (Gare Routière
    Roissypole). You can book nearby hotels such as the Paris Marriott Charles
    de Gaulle Airport Hotel or Holiday Inn Express Paris - CDG Airport."
} 
\end{Verbatim}
\end{tcolorbox}
\caption{\textbf{Qualitative Example of GOATBench data.}}
\label{fig:goatbench_example}
\end{figure*}
\section{GOATBench}
\label{goatbench}

\textbf{GOATBench} is a human-verified benchmark built on top of the GOAT framework.  
It consists of 660 goal-oriented API execution tasks, where solving each task requires planning and invoking a sequence of interconnected APIs.  
Among them, 360 tasks belong to the \textit{seen} category and 300 to the \textit{unseen} category, enabling evaluation across both familiar and novel API compositions.  
While the \textit{seen} set serves as the primary evaluation target, the \textit{unseen} set provides an additional measure of generalization to previously unseen APIs.  
This benchmark was constructed with the GOAT pipeline and subsequently verified and annotated by human experts.  
Qualitative examples are shown in~\cref{fig:goatbench_example}.

The benchmark covers 293 tools and 830 APIs across four user-centric domains—\textit{financial}, \textit{food}, \textit{entertainment}, and \textit{travel}—with all APIs collected from RapidAPI Hub via StableToolBench~\cite{Qin2023ToolLLMFL}.  
RapidAPI provides a hierarchical structure where each tool contains multiple APIs.  
In addition, we implemented two global tools—\textit{compare} and \textit{difference}—that apply across all domains.  
The \textit{compare} tool evaluates numerical similarity or proximity between values, while the \textit{difference} tool determines which value is greater, enabling comparison-based reasoning.  
All LLM-based construction stages use the LLaMA-3-70B-Instruct model~\cite{llama3-70b-instruct}, ensuring reproducibility and easy extension to other API sets.
  
The \textit{seen test} set is created by sampling subgraphs from the same API dependency graphs as training, while modifying parameter values to ensure semantic variation under identical structures and call sequences.  
This set is the main benchmark for measuring performance in in-domain settings.  
The \textit{unseen test} set includes tasks involving tools absent from training, serving as a secondary evaluation to test generalization.
Following~\cite{Qin2023ToolLLMFL}, we also categorize tasks into two types:  
\textbf{Single Tool}, where multiple APIs under the same tool (i.e., a collection of related APIs from a single RapidAPI service) are composed to solve the task; and  
\textbf{Inter Tool}, which requires composing APIs across different tools, testing the agent’s ability to reason over heterogeneous services and handle tool chaining.

A detailed statistical summary of GOATBench, RestBench, and API-Bank is provided in~\cref{tab:benchmark_stats}, including the number of domains, APIs, instances, and average dependency-chain lengths. This comparison highlights the complexity and coverage of GOATBench relative to prior benchmarks. 
\cref{tab:subgraph_sizes} further reports the distribution of subgraph sizes used to construct GOATBench tasks, illustrating the prevalence of 2–4 step dependency chains in real API compositions.

\begin{table}[t]
\centering
\scalebox{0.7}{
\begin{tabular}{lccccccc}
    \toprule
    &
    & \multicolumn{3}{c}{Inter Tool} 
    & \multicolumn{3}{c}{Single Tool} \\
    
    \cmidrule(lr){3-5} \cmidrule(lr){6-8}
    Prompting Method & FT
    & SA & IA & SR 
    & SA & IA & SR \\
    \specialrule{1pt}{2pt}{2pt}
    \multirow{2}{*}{ReACT}
            & \xmark & 7.9 & 1.3 & 5.8 & 19.8 & 1.5  & 8.5 \\
            & \cmark & \textbf{57.6} & \textbf{15.3}  & \textbf{6.7} & \textbf{64.8} & \textbf{23.2} & \textbf{10.5} \\
    \midrule
    \multirow{2}{*}{ReACT + ID}
            & \xmark & 20.6 & 4.7 & 10.1 & 33.3 & 7.3 & 4.0 \\
            & \cmark & \textbf{35.8} & \textbf{19.8} & \textbf{7.2} & \textbf{42.2} & \textbf{31.1} & \textbf{5.9} \\
    \specialrule{1pt}{2pt}{2pt}
    \multirow{2}{*}{Global \textbf{(Baseline)}}       
            & \xmark & 9.7 & 2.9 & 7.2 & 18.2 & 6.6 & 7.9 \\
            & \cmark & \textbf{59.4} & \textbf{26.5} & \textbf{12.3} & \textbf{69.1} & \textbf{33.8} & \textbf{16.5} \\
    \midrule
    \multirow{2}{*}{Global + ID}
            & \xmark & 19.2 & 12.6 & \textbf{8.7} & 37.1 & 21.4 & 9.9 \\
            & \cmark & \textbf{52.8} & \textbf{24.7} & 7.2 & \textbf{68.7} & \textbf{35.3} & \textbf{12.5} \\
    \bottomrule
\end{tabular}
}
\caption{\textbf{Experiment results on seen test set.} ID: Instruction Decomposer. \textbf{GOAT FT}: GOAT fine-tuned. We use Llama-3-8B-Instruct as backbone model. Metrics are SA (Selection Accuracy), IA (Invocation Accuracy), and SR (Success Rate). } 
\label{tab:main_seen}
\end{table}

\begin{table}[t]
\centering
\scalebox{0.8}{
\begin{tabular}{lcccccccc}
    \toprule
    \multirow{2}{*}{Backbone} 
    & \multirow{2}{*}{FT} 
    & \multicolumn{3}{c}{Inter Tool} 
    & \multicolumn{3}{c}{Single Tool} \\
    \cmidrule(lr){3-5} \cmidrule(lr){6-8}
    & & SA & IA & SR & SA & IA & SR \\
    \midrule
    Qwen2-7B    & \xmark & 9.4 & 3.9  & 3.8  & \textbf{23.8} & 6.2  & 7.2 \\
    Qwen2-7B    & \cmark & \textbf{22.4} & \textbf{6.5} & \textbf{4.3}  & 23.6 & \textbf{6.9} & \textbf{8.6} \\
    \midrule
    Llama3-8B   & \xmark & 9.7 & 2.9  & 7.2   & 18.2 & 6.6  & 7.9 \\
    Llama3-8B   & \cmark & \textbf{59.4} & \textbf{26.5} & \textbf{12.3}  & \textbf{69.1} & \textbf{33.8} & \textbf{16.5} \\
    \midrule
    Llama3-70B  & \xmark & 15.3 & 4.5  & 13.0  & 33.9 & 9.0 & 17.1 \\
    Llama3-70B  & \cmark & \textbf{59.5} & \textbf{21.6} & \textbf{14.0}  & \textbf{71.8} & \textbf{28.9} & \textbf{30.9} \\
    \bottomrule
\end{tabular}
}
\caption{\textbf{Comparison across backbone models on GOATBench.} We use \textit{Baseline} as the prompting method.}
\label{tab:multi_model_comparison}
\end{table}
\begin{table}[t]
\centering
\scalebox{0.8}{
\begin{tabular}{lccc}
\toprule
\multicolumn{1}{c}{} & 
\multicolumn{1}{c}{} & 
\multicolumn{1}{c}{TMDB} & 
\multicolumn{1}{c}{Spotify} \\
\cmidrule(lr){1-1} \cmidrule(lr){2-2} \cmidrule(lr){3-3} \cmidrule(lr){4-4}
Backbone & FT & Success (\%) & Success (\%) \\
\midrule
Llama3-8B & \xmark & 1  & 10.5 \\
Llama3-8B & \cmark & \textbf{12} & \textbf{21.1} \\
Qwen2.5-7B & \xmark & 0  & 8.8 \\
Qwen2.5-7B & \cmark & \textbf{6} & \textbf{29.8} \\
\bottomrule
\end{tabular}
}
\caption{\textbf{Comparison across backbone models on RestBench.} We use \textit{Baseline} as the prompting method.}
\label{tab:restbench_multi_model_comparison}
\end{table}

\begin{table}[t]
\centering
\scalebox{0.8}{
\begin{tabular}{lcc}
\toprule
Backbone & Training Method & Success (\%) \\
\midrule
Llama3-8B & \xmark & 34 \\
Llama3-8B & \cmark & \textbf{64} \\
Qwen2.5-7B & \xmark & 28 \\
Qwen2.5-7B & \cmark & \textbf{66} \\
\bottomrule
\end{tabular}
}
\caption{\textbf{Comparison across backbone models on APIBank(Plan+Retrieve+Call).} We use \textit{Baseline} as the prompting method.}
\label{tab:apibank_multi_model_comparison}
\end{table}
\section{LLM Agent Prompting Baselines}
\label{appendix_llmagent}

We test our approach using following four LLM agent prompting baaselines designed to perform goal-oriented tasks. Each LLM agent design consists of four key components: API retrieval, API selection, API call generation, and final response generation. The agent designs differ in how and at what level the LLM performs planning.

\noindent\textbf{ReACT}~\citep{yao2023react} \ \ 
ReACT is a reflective agent that performs planning in an iterative manner.
Since a goal-oriented user query often requires a sequence of dependent API calls, ReACT can serve as the baseline planning method.
At each timestep, it jointly selects the next API to call and generates its arguments, conditioned on the full history of prior API calls and their outputs.
Given the user query $u$, $k$ potentially relevant documents are first retrieved using an external retriever.
The same set of the retrieved documents are then fed to the agent at each timestep.

\noindent\textbf{Global Planner}~\citep{huang2024understandingplanningllmagents} \ \ 
Given $k$ relevant documents retrieved as in ReACT, this agent performs global planning by determining the entire sequence of API functions with a single prompt, using only the user query and the retrieved API documents. 
This global planning strategy enables the agent to optimize the overall call sequence from a holistic perspective.
To reflect the outputs of the previous API calls,
we iteratively generate the API calls based on the planned API function sequence.
Then, API calls are executed iteratively, incorporating the previously executed API outputs.

\noindent\textbf{ReACT + Instruction Decomposer} \ \ 
This agent extends ReACT by incorporating iterative planning at the natural language level. 
Starting from the user query, it generates one subinstruction at a time, conditioned on the original query, prior subinstructions, API calls, and their outputs. 
For each subinstruction, it retrieves the $k$ most relevant API document and jointly performs API selection and call generation.
This process continues step-by-step, with each execution result guiding the generation of the next subinstruction.

\noindent\textbf{Global Planner + Instruction Decomposer} \  \
This agent also performs language-based subinstruction planning but returns global plans in a single shot.
For each subinstruction, it then retrieves $k$ relevant API documents, and jointly performs API selection and call generation in the given subinstruction order, reflecting the planned execution flow.

\cref{tab:main_seen} compares the results of applying GOAT fine-tuning across different prompting methods.
We observe consistent performance improvements over all methods, demonstrating the broad effectiveness of our approach.
Among them, the decomposition-first method with a global planning strategy achieves the largest gains, and we therefore adopt it as our primary baseline for the main experiments.

\begin{figure*}[t]
\begin{tcolorbox}[colframe=gray!50!black, colback=gray!10!white, title=Zero-shot Inference Example, width=\textwidth, sharp corners=southwest]
\scriptsize
\begin{Verbatim}
{
    "user query": "What is the birthday of the most popular person right now?",
    "api_path": [],
    "final_response": "I apologize, but the API did not provide any data on the most 
    popular person. Please provide a valid API response or clarify the query to get
    a more accurate answer."
}
\end{Verbatim}
\end{tcolorbox}
\begin{tcolorbox}[colframe=gray!50!black, colback=gray!10!white, title=GOAT Fine-tuned Inference Example, width=\textwidth, sharp corners=southwest]
\scriptsize
\begin{Verbatim}
 {
     "user query": "What is the birthday of the most popular person right now?",
    "api_calls": [
        {
            "api_name": "GET /person/popular",
            "input": {
                "page": 1
            },
            "output": {
              "id": 53,
              "name": "Peter Stormare",
              "original_name": "Peter Stormare",
              ...
            },
        {
            "api_name": GET /person/{person_id},
            "input": {
                "person_id": "53"
            },
            "output": {
              "id": 53,
              "name": "Peter Stormare",
              "birthday": "1953-08-27",
              ...
            }

        }
    ],
    "final_response": "The most popular person is Peter Stormare, a Swedish actor
    born on August 27.",
}
\end{Verbatim}
\end{tcolorbox}
\caption{\textbf{Comparison of zero-shot inference result and GOAT fine-tuned inference result on RestBench.}}
\label{fig:qual_ex_1}
\end{figure*}
\begin{figure*}[t]
\begin{tcolorbox}[colframe=gray!50!black, colback=gray!10!white, title=Zero-shot Inference Example, width=\textwidth, sharp corners=southwest]
\scriptsize
\begin{Verbatim}
{
    query": "Retrieve a list of recommended movies based on Peter's preferences 
    and filter out the movies that the user has already watched.",
    "api_path": [
        {
            "api_name": "UserMoviePreferences",
            "input": {
                "user_name": "Peter"
            },
            "output": {
                "preferences": ["Action", "Drama","Thriller"]
            }
        },
        {
            "api_name": "UserWatchedMovies",
            "input": {
                "user_name": "Peter"
            },
            "output": ["The Matrix", "The Lord of the Rings","The Dark Knight"]
        },
        {
            "api_name": "MovieRecommendations",
            "input": {},
            "output": {"error": "missing 1 required positional argument: 'preferences'"}
        }
        ],
        "final_response": "Based on Peter's preferences for Action, Drama, and Thriller, 
        here are some recommended movies: The Shawshank Redemption, 
        The Silence of the Lambs, The Dark Knight Rises."
}
\end{Verbatim}
\end{tcolorbox}
\begin{tcolorbox}[colframe=gray!50!black, colback=gray!10!white, title=GOAT Fine-tuned Inference Example, width=\textwidth, sharp corners=southwest]
\scriptsize
\begin{Verbatim}
{
    "query": "Retrieve a list of recommended movies based on Peter's preferences 
    and filter out the movies that the user has already watched.",
    "api_path": [
        {
            "api_name": "UserMoviePreferences",
            "input": {
                "user_name": "Peter"
            },
            "output": {
                "preferences": ["Action", "Drama","Thriller"]
            }
        },
        {
            "api_name": "UserWatchedMovies",
            "input": {
                "user_name": "Peter"
            },
            "output": ["The Matrix", "The Lord of the Rings","The Dark Knight"]
        },
        {
            "api_name": "MovieRecommendations",
            "input": {
                "preferences": ["Action", "Drama","Thriller"]
            },
            "output": {
                "recommended_movies": [
                    "The Dark Knight",
                    "The Lord of the Rings",
                    "The Shawshank Redemption",
                    "Forrest Gump",
                    ...
    ],
    "final_response": "Based on your preferences for Action, Drama, and Thriller movies,
    I recommend the following movies that you haven't watched yet: 
    The Shawshank Redemption, Forrest Gump, Joker, Inception, Parasite, and Get Out."
}
\end{Verbatim}
\end{tcolorbox}
\caption{\textbf{Comparison of zero-shot inference result and GOAT fine-tuned inference result on API-Bank.}
}
\label{fig:qual_ex_2}
\end{figure*}
\begin{figure*}[t]
\begin{tcolorbox}[colframe=gray!50!black, colback=gray!10!white, title=Zero-shot Inference Example, width=\textwidth, sharp corners=southwest]
\scriptsize
\begin{Verbatim}
{
    "query": "Get the detailed information from the search results of books with 
    the name 'harry potter'.",
    "api_path": [],
    "final_response": "No search results found for 'Harry Potter'."
}
\end{Verbatim}
\end{tcolorbox}
\begin{tcolorbox}[colframe=gray!50!black, colback=gray!10!white, title=GOAT Fine-tuned Inference Example, width=\textwidth, sharp corners=southwest]
\scriptsize
\begin{Verbatim}
{
    "query": "Get the detailed information from the search results of books with 
    the name 'harry potter'.",
    "api_path": [
        {
            "api_name": "HAPI Books/Search Books by Name",
            "input": {
                "book_name": "harry+potter"
            },
            "output": [
              {
                "book_id": 136251,
                "name": "Harry Potter and the Deathly Hallows",
                ...
              },
              {
                "book_id": 42844155,
                "name": "Harry Potter and the Sorcerer’s Stone",
                ...
              },
              ...
            ]
        },
        {
            "api_name": "HAPI Books/Get Book Information by Book Id",
            "input": {
                "book_Id": "136251"
            },
            "output": {
              "name": "Harry Potter #7Harry Potter and the Deathly Hallows",
              "authors": ["J.K. Rowling"],
              "rating": 4.62,
              "pages": 759,
              "published_date": "July 21, 2007",
              "synopsis": "Harry has been burdened with a seemingly impossible task: 
              that of locating and destroying Voldemort's remaining Horcruxes. ..."
            }
        }
    ],
    "final_response": "Here are the results of the search for books with the name
    'Harry Potter': Harry Potter and the Deathly Hallows by J.K. Rowling, published in 
    2007, has a rating of 4.62. Here is a summary of Harry Potter and the Deathly Hallows: 
    Harry has been tasked with locating and destroying Voldemort's remaining Horcruxes,
    ... "
}
\end{Verbatim}
\end{tcolorbox}
\caption{\textbf{Comparison of zero-shot inference result and GOAT fine-tuned inference result on GOATBench.}}
\label{fig:qual_ex_3}
\end{figure*}
\section{Qualitative Results}
Qualitative result examples on each benchmark are in~\cref{fig:qual_ex_1,fig:qual_ex_2,fig:qual_ex_3}.

\section{Effects of GOAT with Diverse LLMs}
\label{appendix_resultsllm}
\cref{tab:multi_model_comparison,tab:restbench_multi_model_comparison,tab:apibank_multi_model_comparison} further confirms that GOAT consistently boosts performance across different LLM backbones, including Qwen2-7B-Instruct, Qwen2.5-7B-Instruct, Llama3-8B-Instruct, and Llama3-70B-Instruct. These results highlight the robustness and generality of our fine-tuning strategy.
Notably, in \cref{tab:multi_model_comparison}, even for Llama3-70B, where the same model was used both to generate training data and to perform fine-tuning, we still observe substantial gains.
As discussed in~\cref{discussion}, this is because our call-first generation strategy provides reliable supervision: by starting from executable API paths and asking the model only to abstract them into natural-language queries, we avoid pitfalls of instruction-first methods and enable the model to benefit from supervision signals even when self-generated.

\section{Results on Unseen Test}
\label{appendix_resultsunseen}
\begin{table}[t]
\centering
\scalebox{0.85}{
\begin{tabular}{ccccccc}
    \toprule
    \multirow{2}{*}{GOAT FT} 
    & \multicolumn{3}{c}{Inter Tool} 
    & \multicolumn{3}{c}{Single Tool} \\
    
    \cmidrule(lr){2-4} \cmidrule(lr){5-7}
    & SA & IA & SR 
    & SA & IA & SR \\
    \specialrule{1pt}{2pt}{2pt}
    \xmark & 15.5 & 7.7 & 13.5 & 18.5 & 10.5 & 21.2 \\
    \cmark & \textbf{42.5} & \textbf{19.8} & \textbf{21.2} & \textbf{42.3} & \textbf{23.5} & \textbf{28.8} \\
    \bottomrule
\end{tabular}
}
\caption{\textbf{Experiment results on unseen test set.} We use Llama3-8B-Instruct backbone and \textit{Baseline} prompting method.}
\label{tab:main_unseen}
\end{table}

\begin{table}[t]
\centering
\scalebox{0.85}{
\begin{tabular}{ccc}
    \toprule
    Training Method & TMDB & Spotify \\
    \specialrule{1pt}{2pt}{2pt}
    Zero-shot & 1.0 & 10.5 \\
    GOAT (GOATBench) & 10.0 & 14.0 \\
    GOAT (RestBench) & \textbf{12.0} & \textbf{21.1} \\
    \bottomrule
\end{tabular}
}
\caption{\textbf{Cross-benchmark generalization on RestBench.} All experiments use the Llama3-8B-Instruct backbone with the \textit{Baseline} prompting method.}
\label{tab:restbench_generalization}
\end{table}
Although the unseen setting is not the primary target of our benchmark, we observe that models trained on GOAT still exhibit strong generalization performance. As shown in Table~\ref{tab:main_unseen}, fine-tuning both the LLM and retriever leads to consistent improvements across all task types. This demonstrates that task-aligned fine-tuning not only benefits in-domain execution but also improves robustness to previously unseen API combinations, though the gains are more limited compared to the seen setting.

In addition to the unseen split within GOATBench, we further evaluate cross-benchmark generalization by training models on GOATBench and directly testing them on RestBench, which consists of entirely disjoint APIs and domains (e.g., TMDB, Spotify). This setting provides a more stringent evaluation of out-of-domain generalization, as there is no overlap in API schemas or domain-specific knowledge between training and testing.
As shown in Table~\ref{tab:restbench_generalization}, models trained on GOATBench achieve consistent improvements over the zero-shot baseline across both domains. These results suggest that GOAT training helps models acquire transferable agentic reasoning skills that generalize beyond the training APIs.

\begin{table}[t]
\centering
\scalebox{0.9}{
\begin{tabular}{lc}
    \toprule
    Data Generation Method & Success (\%) \\
    \specialrule{1pt}{2pt}{2pt}
    Zero-shot & 34 \\
    ToolLLM & 44 \\
    GOAT & \textbf{64} \\
    \bottomrule
\end{tabular}
}
\caption{\textbf{Data generation method ablation results on API-Bank (Plan+Retrieve Call).} All experiments use the \textit{Baseline} prompting method with a Llama-3-8B backbone.}
\label{tab:apibank_comparison}
\end{table}

\begin{table}[t]
\centering
\scalebox{0.9}{
\begin{tabular}{lcc}
    \toprule
    Data Generation Method & TMDB & Spotify \\
    \cmidrule(lr){2-3}
     & Success (\%) & Success (\%) \\
    \specialrule{1pt}{2pt}{2pt}
    Zero-shot & 1 & 10.5 \\
    ToolLLM & 5 & 10.5 \\
    GOAT (finetuned) & \textbf{12} & \textbf{21.1} \\
    \bottomrule
\end{tabular}
}
\caption{\textbf{Data generation method ablation results  on RestBench.} All experiments use the \textit{Baseline} prompting method with a Llama-3-8B backbone.}
\label{tab:restbench_generalization}
\end{table}
\section{Ablation Results on Training Data Generation Methods}
To further validate the generalization capability of GOAT, we conduct additional evaluations on external benchmarks and compare our method with ToolLLM, a representative data generation framework for tool-augmented agents.
On API-Bank (Table~\ref{tab:apibank_comparison}), both ToolLLM and GOAT improve over the zero-shot baseline, but GOAT achieves substantially higher performance (44\% $\rightarrow$ 64\%). On RestBench (Table~\ref{tab:restbench_generalization}), while ToolLLM provides only marginal gains over the baseline (e.g., 1\% $\rightarrow$ 5\% on TMDB), GOAT consistently yields larger improvements (up to 12\% on TMDB and 21.1\% on Spotify), indicating its superior ability to learn transferable, multi-step reasoning for tool use in unseen environments. It is also worth noting that ToolLLM training data is generated using GPT-3.5-Turbo following the original setup, while GOAT uses synthetic data generated with a smaller Llama-3-70B-Instruct model. Despite this, GOAT consistently outperforms ToolLLM across both benchmarks, further emphasizing the effectiveness of our framework.

\begin{table}[t]
\centering
\scalebox{0.8}{
\begin{tabular}{lcccccc}
    \toprule
    \multirow{2}{*}{LLM FT Method} 
    & \multicolumn{3}{c}{Inter Tool} 
    & \multicolumn{3}{c}{Single Tool} \\
    
    \cmidrule(lr){2-4} \cmidrule(lr){5-7}
    & SA & IA & SR 
    & SA & IA & SR \\
    \specialrule{1pt}{2pt}{2pt}
    LoRA                  & 44.0 & 19.2 & 12.0 & 51.3 & 31.7 & 9.9 \\
    LoRA + Self-Distill   & 45.5 & 18.5 & 11.5 & 52.8 & 33.0 & 13.8 \\
    LoRA + Masking Args   & \textbf{59.4} & \textbf{26.5} & \textbf{12.3} & \textbf{69.1} & \textbf{33.8} & \textbf{16.5} \\
    \bottomrule
\end{tabular}
}
\caption{\textbf{Ablation results on different LLM fine-tuning methods with Global design.} We use Llama-3-8B-Instruct model as backbone. All models use finetuned SBERT as the retriever.}
\label{tab:ftmethod_ablation}
\end{table}
\section{Ablation Results on LLM Fine-tuning Methods}

\cref{tab:ftmethod_ablation} compares different fine-tuning strategies designed to mitigate overfitting in the LLM. We observed that standard LoRA-based fine-tuning tends to cause the model to overfit to specific argument values seen during training, leading to reduced generalization. To address this issue, we explored two approaches: (1) self-distillation with soft targets from a pretrained model, and (2) masking argument tokens by setting their loss contributions to zero. 
Our results show that the masking strategy is particularly effective, as it prevents memorization of argument values and encourages structural learning of API call formats.

\begin{table}[t]
\centering
\scalebox{0.9}{
\begin{tabular}{ccc}
    \toprule
    GOAT FT & Recall@GT & Recall@5 \\
    \midrule
    \xmark & 19.5 & 30.0 \\
    \cmark & \textbf{63.1} & \textbf{81.5} \\
    \bottomrule
\end{tabular}
}
\caption{\textbf{Retriever performance on GOATBench.} We use the all-MiniLM-L6-v2 encoder as the retriever. Recall@GT retrieves the same number of documents as the ground-truth API calls for each example.}
\label{tab:retriever-results}
\end{table}
\section{Ablation Results on API Document Retriever}

We conduct an ablation study to evaluate the retriever performance in isolation using recall metrics on GOATBench. \cref{tab:retriever-results} reports Recall@GT and Recall@5 when retrieving relevant API documents given the user query $u$. Recall measures how many of the retrieved documents are actually correct. Recall@GT evaluates this by retrieving the same number of documents as the ground-truth API calls for each data point.
The results show clear performance improvements through fine-tuning.



\section{Data Generation Prompts}
\subsection{API Document Parsing Prompt}
\label{parsing prompt}
See~\cref{fig:doc_parsing} for an example prompt.

\subsection{LLM Filtering Prompt}
\label{llm filtering prompt}
See~\cref{fig:LLMfiltering-prompt} for an example prompt.

\subsection{Actual Call Output Filtering Prompts}
\label{api call filtering prompt}
See~\cref{fig:APICalling-prompt},~\cref{fig:APIfiltering-prompt} for an example prompt.

\subsection{API Call Sequence Generation Prompt}
\label{api call sequence generation prompt}
See~\cref{fig:APIpath-prompt} for an example prompt.

\subsection{Sub-instruction Generation Prompt}
\label{subinstruction generation prompt}
See~\cref{fig:subinst-prompt} for an example prompt.

\subsection{User Query Generation}
\label{query generation}
See~\cref{fig:query-prompt} for an example prompt.

\subsection{Final Response Generation}
\label{final response generation}
See~\cref{fig:finalresponse-prompt} for an example prompt.

\section{Success Rate Prompt}
See~\cref{fig:Success Rate Prompt} for prompt used in SR evaluation on GOATBench.

\section{Use of LLMs}
We acknowledge that LLMs were used as writing assistants to improve grammar, clarity, and readability of the manuscript.

\begin{figure*}[t]
\begin{tcolorbox}[colframe=gray!50!black, colback=gray!10!white, title=API Document Parsing Prompt, width=\textwidth, sharp corners=southwest]
\scriptsize
\begin{verbatim}
You are an API Documentation Assistant responsible for analyzing API documentation
and summarizing the semantics of each input parameter and the output of the API function.

You will be provided with:
1. API Document: A dictionary containing information about an API function, with details.

Your task is to:
1. Provide a clear semantic description of what each input parameter 
and output of the API function represents.
2. There can be multiple input parameters, including both required and optional 
parameters.
3. If there are no required or optional parameters, return empty array 
for input parameter description.

Output Format:
- You must return a dictionary with the keys "input_params" and "output".
- "input_params": Return an array of semantic descriptions for each input parameter.
                 If there is None, return empty array.
- "output": Return a semantic description for output of the API function.

ONLY return the dictionary as your output. DO NOT include any other words.
\end{verbatim}
\end{tcolorbox}
\caption{\textbf{Prompt used for API document parsing.}}
\label{fig:parsing_prompt}
\end{figure*}

\begin{figure*}[t]
\begin{tcolorbox}[colframe=gray!50!black, colback=gray!10!white, title=LLM Filtering Prompt, width=\textwidth, sharp corners=southwest]
\scriptsize
\begin{verbatim}
You are an API Documentation Assistant responsible for determining whether two APIs 
can be connected 
sequentially, i.e. the output of the first API must be used as the input for the 
second API.

You will be provided with:
1. API1 Document: A dictionary containing the details of API1's output.
2. API1 Semantic Descriptions: Natural language explanations of API1's output
3. API2 Document: A dictionary containing the details of API2's input.
4. API2 Semantic Descriptions: Natural language explanations of API2's input.

Your task is to:
1. Analyze the semantic descriptions and the provided API documents to determine if 
API1's output
   can be used as API2's input.
2. Return True only if the information in the output of API1 can be used as a valid 
input for API2.
3. Do not return True when input of API1 can be reused in API2.
4. Explain why the APIs are connectable or not.

Output Format:
- You must return a dictionary with the keys "connectable" and "reason".
- "connectable": Return True only if API1's output can be used as API2's input, 
otherwise return False.
- "reason": Provide a clear explanation describing why the APIs can or cannot be 
connected.

ONLY return the dictionary as your output. DO NOT include any other words.
\end{verbatim}
\end{tcolorbox}
\caption{\textbf{Prompt used for filtering edges via LLM.}}
\label{fig:LLMfiltering-prompt}
\end{figure*}
\begin{figure*}[t]
\begin{tcolorbox}[colframe=gray!50!black, colback=gray!10!white, title=API Call Generating Prompt for Edge Filtering - 1. First Call, width=\textwidth, sharp corners=southwest]
\scriptsize
\begin{verbatim}
You are an API Documentation Assistant responsible for generating function calls
based on API documentation.

You will be provided with:
1. API Document: A dictionary containing information about an API function, with details.

Your task is to:
1. Create a fictional scenario where you need to use the API.
2. Populate the API function's required parameters and optional parameters with 
appropriate values, ensuring that all required parameters are included and match the 
correct data types.

Output Format:
- You must return a dictionary where each parameter name is the key, and the parameter 
  value is the value of the dictionary.
- Ensure each parameter value has the correct data types.
- If there are no required or optional parameters for the API function, return an empty 
  dictionary.

ONLY return the parameter dictionary as your output. DO NOT include any other words.
\end{verbatim}
\end{tcolorbox}
\begin{tcolorbox}[colframe=gray!50!black, colback=gray!10!white, title=API Call Generating Prompt for Edge Filtering - 2. Subsequent Call, width=\textwidth, sharp corners=southwest]
\scriptsize
\begin{verbatim}
You are an API Documentation Assistant responsible for generating function calls 
based on API documentation and previous API call results.

You will be provided with:
1. API Document: A dictionary containing information about an API function, 
   including parameter names, data types, and descriptions.
2. API Call Results: The result of one or more previous API function calls.
3. Reason: An array explaining how the API Call Results can be used to populate 
   the parameters for the current API call.

Your task is to:
1. Create a fictional scenario where you need to use the API.
2. Populate the API function’s required and optional parameters using the following rules:
   - First, use values justified by the API Call Results and the Reason array.
   - If a parameter cannot be filled this way, infer it using the information in the API 
   Document 
     (e.g., parameter descriptions or type hints).
3. Ensure all parameter values match the correct data types as specified in the API 
   Document.

Output Format:
- Return a dictionary where each key is a parameter name and the value is the parameter’s 
  value.
- If no parameters can be populated from the available information, return an empty 
  dictionary.

ONLY return the parameter dictionary as your output. DO NOT include any other text.
\end{verbatim}
\end{tcolorbox}
\caption{\textbf{Prompt used for API call generation for each edges.}}
\label{fig:APICalling-prompt}
\end{figure*}
\begin{figure*}[t]
\begin{tcolorbox}[colframe=gray!50!black, colback=gray!10!white, title=API Call Filtering Prompt, width=\textwidth, sharp corners=southwest]
\scriptsize
\begin{verbatim}
You are an API Documentation Assistant responsible for determining if the information from 
the result of the first API call is used in the parameters of the second API call.
You will be provided with:
1. api_result: A result from the first API call.
2. llm_result: Parameters and their values for calling next API.

Your task is to:
1. Analyze the contents of api_result to determine if it was used as input in llm_result.
2. Provide an explanation about whether or not the first API result influenced the 
  parameters of the next API call.

Output Format:
- You must return a dictionary with the keys "connectable" and "reason".
- "connectable": Return True if api_result was used in llm_result, otherwise return False.
- "reason": Provide a clear explanation describing why api_result was or was not used 
  as part of llm_result.

ONLY return the dictionary as your output. DO NOT include any other words.
\end{verbatim}
\end{tcolorbox}
\caption{\textbf{Prompt used for filtering edges via API Call Output.}}
\label{fig:APIfiltering-prompt}
\end{figure*}

\begin{figure*}[t]
\begin{tcolorbox}[
colframe=gray!50!black, colback=gray!10!white, title=Make First Call, width=\textwidth, sharp corners=southwest]
\scriptsize
\begin{verbatim}
You are an API Documentation Assistant responsible for constructing parameter values 
for API calls based on API documentation.
You will be provided with:
1. API Document: A dictionary containing information about an API function, with details.
Your task is to:
1. Create a fictional scenario where you need to use the API.
2. Populate the API function's required parameters and optional parameters 
   with appropriate values, ensuring that all required parameters are included 
   and match the correct data types.
Output Format:
- Return a dictionary where each parameter name is the key, and the parameter value is 
  the value of the dictionary.
- Ensure each parameter value has the correct data types.
- If there are no required or optional parameters for the API function, return an empty 
  dictionary.
ONLY return the parameter dictionary as your output. DO NOT include any other words.
\end{verbatim}
\end{tcolorbox}
\end{figure*}

\begin{figure*}[t]
\begin{tcolorbox}[colframe=gray!50!black, colback=gray!10!white, title=Make Call Step 1, width=\textwidth, sharp corners=southwest]
\scriptsize
\begin{verbatim}
You are an API Documentation Assistant responsible for constructing parameter values 
for API calls based on API documentation and previous API call results.
You will be provided with a dictionary containing the following keys:
1. `API Document`:
   This key provides information about an API function, including its details. 
   It should be used solely to understand the API and identify its required and optional 
   parameters.
   - **Important:** Do not use any values from the `API Document` directly to populate 
   parameters
   for the API call.
2. `Parameter Dictionary`:
   This key contains a dictionary where each key is a parameter index, and each value is 
   the corresponding parameter name. This is used to reference parameters by their 
   indices.
3. `Parameter Value`:
   This key contains a dictionary that maps each parameter index to a dictionary detailing 
   how to obtain the parameter's value based on previous API call results:
   - Each value includes:
     - `docid`: The unique ID of the document from which the parameter value is derived. 
     This `docid` corresponds directly to a `docid` in the `Previous Result`, indicating 
     the source of the data to be used.
     - `reason`: A brief explanation of how the specific data from the previous results 
     (API1) 
     is suitable to be used as a parameter in the current API call (API2).
4. `Previous Result`:
   This key contains a dictionary of results from previous API function calls. 
   Each key is a `docid` that corresponds to a previous API call, and each value contains 
   the results returned by that call. The `docid` used here matches the `docid` referenced 
   in the `Parameter Value`.
### Your task is to follow these steps:
1. **Identify Parameter Names**:
   - Use the `Parameter Dictionary` to reference the names of parameters using their 
   indices 
   provided in the `Parameter Value`.
2. **Extract Parameter Values**:
   - For each parameter identified, use its index to find the corresponding `docid` 
   and `reason` in the `Parameter Value`.
   - Locate the specific data in `Previous Result` based on the `docid` and ensure 
   the data matches the reasons and conditions for use.
   - The results from `Previous Result` (API1) will be applied to the parameters in 
   the current API call (API2) following the explanations in the `reason`.
3. **Populate the Dictionary**:
   - Create a dictionary where each parameter name (from the `Parameter Dictionary`) is 
   the key, and the extracted value from `Previous Result` is the corresponding value.
   - Populate only those parameters that are explicitly mentioned in the 
   `Parameter Value`. 
   Exclude all others.
   - **DO NOT use any default values or other values from the `API Document` to populate
   parameters.**
4. **Validate and Output**:
   - Confirm that all parameters listed in the `Parameter Value` are properly populated 
   without using default or unrelated values from the `API Document`.
   - Return a dictionary where each parameter name is the key and the parameter value is 
   the value of the dictionary.
   - If no parameters can be properly populated using the provided data and reasons, 
   return an empty dictionary.
ONLY return the parameter dictionary as your output. DO NOT include any other words.
\end{verbatim}
\end{tcolorbox}
\end{figure*}

\begin{figure*}[t]
\begin{tcolorbox}[colframe=gray!50!black, colback=gray!10!white, title=Make Call Step 2, width=\textwidth, sharp corners=southwest]
\scriptsize
\begin{verbatim}
You are an API Documentation Assistant responsible for completing function call parameters 
based on the API documentation and a partially filled parameter dictionary.
You will be provided with:
1. `API Document`: A dictionary containing information about the API function, including 
   its details, required parameters, optional parameters, and their respective default 
   values.
2. `Partially Filled Parameters`: A dictionary where some parameters have already been
   populated, but others are still missing.
Your task is to:
1. Review the `API Document` to identify which parameters (required and optional) are 
   still missing from the `Partially Filled Parameters` dictionary.
2. Populate the missing parameters based on the following rules:
   - Fill in missing parameters with appropriate values that align with the parameter 
     descriptions in the `API Document`. Use your judgment to select realistic and 
     suitable values.
   - Ensure all required parameters are included with appropriate values.
   - Optional parameters can remain unfilled if no suitable value can be determined.
3. Ensure that all parameter values match the correct data types specified in the 
   `API Document`.
Output Format:
- Return a dictionary where each parameter name is the key, and the parameter value is 
  the value of the dictionary.
- The dictionary must include all required parameters (filled with appropriate values) 
  and may include optional parameters (if filled).
- Do not include any other words or explanations in the output.
ONLY return the completed parameter dictionary as your output.
\end{verbatim}
\end{tcolorbox}
\caption{\textbf{API call sequence generation prompt.}}
\label{fig:APIpath-prompt}
\end{figure*}

\begin{figure*}[t]
\begin{tcolorbox}[colframe=gray!50!black, colback=gray!10!white, title=Sub-instruction Prompt, width=\textwidth, sharp corners=southwest]
\scriptsize
\begin{verbatim}
You are an instruction generation assistant for generating lanuage instruction 
that enables execution of given API call.
You will be provided a dictionary containing the following keys:
1. 'API Document': A structured description of the API, including its purpose, 
   required and optional parameters, and any relevant context about its functionality.
2. 'API call': A dictionary of specific parameter values intended for execution of 
   the API call. You must generate language instruction that enables execution of 
   this call.
3. 'Previous API Response': The output or result from preceding API calls. 
   Some values in 'API call' references values in this result. If this is empty, 
   it should not be referenced.
### Your task is to follow these steps:
1. ** Classify Parameters in 'API call':
    - For each key in 'API call', check if its value can be directly derived from 
      the 'Previous API Response'.
    - Classify keys into two groups:
        a. Derived Parameters: Parameters whose values are obtained from 
        the 'Previous API Response'.
        b. Fixed Parameters: Parameters with values that are not contained 
        in 'Previous API Response'.
2. ** Generate Language Instruction:
    - Generate a clear and concise language instruction that enables the execution of 
      the 'API call'.
    - Use the 'API Document' to understand the intent of the 'API call' and ensure that 
      the generated instruction aligns with its goal. The instruction must be 
      goal-oriented, actionable, and contextually accurate.
    - Incorporate the parameter classification:
        a. Derived Parameters: For parameters classified as derived, include in 
           the instruction a detailed explanation of how their values are obtained from 
           the 'Previous API Response.' Clearly reference the specific part or context of 
           the 'Previous API Response' used to derive these values.
           Do not include the derived value itself in the instruction. Instead, 
           describe the reasoning behind its selection, such as it being the first item, 
           the most recent value, the largest attribute, or another logical criterion. 
           The reasoning must be explicit and actionable.
        b. Fixed Parameters: For parameters classified as fixed, include their 
           specific values directly in the instruction. Ensure these values are 
           explicitly stated to avoid ambiguity.
    - The instruction should naturally integrate both types of parameters and describe 
      the action to be performed in a clear and executable manner.
Output Format:
- Return a dictionary with the following keys:
    - "thought": Provide a brief but clear explanation of your reasoning process, 
    including how parameters were classified, how derived values were selected, 
    and how they were incorporated into the instruction.
    - "instruction": Generate a concise, goal-oriented sentence that describes 
    the action required to execute the given API call. Ensure the instruction 
    integrates both derived and fixed parameters appropriately, specifying 
    derived parameter contexts, the reasoning for their selection, and explicitly 
    stating fixed parameter values.
DO NOT use the vague terms such as "use the obtained value" or "from specific values."
DO NOT include parameter names or technical jargon from the API. 
Translate these into natural language descriptions of their role or value.
ONLY return the output dictionary. DO NOT include any other words.
\end{verbatim}
\end{tcolorbox}
\caption{\textbf{Sub-instruction generation prompt.}}
\label{fig:subinst-prompt}
\end{figure*}
\begin{figure*}[t]
\begin{tcolorbox}[colframe=gray!50!black, colback=gray!10!white, title=Query Prompt, width=\textwidth, sharp corners=southwest]
\scriptsize
\begin{verbatim}
You are an API Documentation Assistant. Your role is to interpret a list of 
subinstructions—understanding that each subinstruction is planned, executed, 
and possibly leads to creating or adapting subsequent subinstructions based on 
prior outcomes—and convert them into a single, high-level user query that reflects 
their collective intent without revealing any internal steps or technical API jargon.
### Provided Information:
- Subinstructions: A sequence of iterative steps working toward a single overarching 
  objective. They are planned and executed in order, and each result can influence the 
  creation or modification of the next subinstruction.
### Your Task:
1. Infer the broader purpose by analyzing how these subinstructions connect logically 
   and build upon each other’s results.
2. Synthesize them into one natural, user-friendly query that preserves crucial details 
   and dependencies but does not mention the subinstructions themselves.
3. Represent information at a high level wherever possible, but retain all specific 
   details 
   (e.g., IDs, names, dates) from the **first subinstruction** exactly as they are.
4. For subinstructions after the first one, prioritize connecting them through context 
   (e.g., "first video," "latest episode") rather than using specific identifiers 
   unless absolutely necessary.
5. Ensure that every subinstruction meaningfully contributes to the final query, 
   preventing any extraneous or unaligned steps.
6. Avoid any technical language or references to specific APIs in the final query.
### Guidelines:
- Include all essential identifiers or conditions (e.g., names, dates, relevant context) 
  from the subinstructions. Do not omit or generalize key details from the 
**first subinstruction**.
- For subsequent subinstructions, derive necessary information from the results of prior 
  steps whenever possible. Use contextual references to maintain continuity without  
  repeating specific identifiers unless required.
- Reflect the necessary sequence or dependency implied by the iterative nature of 
  subinstructions without explicitly describing each step.
- The final query must reflect the intent of **all subinstructions** to ensure no step 
  becomes irrelevant or disconnected.
### Output Format:
- Return only a dictionary with two keys:
- "thought": A short explanation of how you derived the final query from the 
  subinstructions.
- "query": The single-sentence user query representing the overall goal.
DO NOT include API names or technical jargon from the API.
ONLY return the dictionary as your output. DO NOT include any other words.
\end{verbatim}
\end{tcolorbox}
\caption{\textbf{User query generation prompt.}}
\label{fig:query-prompt}
\end{figure*}
\begin{figure*}[t]
\begin{tcolorbox}[colframe=gray!50!black, colback=gray!10!white, title=Final Response Prompt, width=\textwidth, sharp corners=southwest]
\scriptsize
\begin{verbatim}
You are an answer generation assistant tasked with providing natural language responses 
to user queries by analyzing API call results.
You will be provided a dictionary containing the following keys:
1. 'User Query': A natural language question or request from the user.
2. 'API Call Result': A list of dictionaries, each representing a step or subinstruction 
   carried out to fulfill the user query. Each dictionary contains:
- 'subinstruction': A brief description of the step taken.
- 'api response': The actual data or result obtained from executing the subinstruction.
### Your task is to follow these steps:
1. ** Analyze API Call Result: **
- Examine each dictionary in the 'API Call Result' list.
- Understand the purpose of each 'subinstruction' and the corresponding 'api response.'
- Identify how each 'api response' contributes to answering the 'User Query.'
- If necessary, combine results from multiple subinstructions to generate a comprehensive 
  answer.
2. ** Generate Final Answer: **
- Construct a coherent and natural response to the 'User Query' based on the collected 
  information from 'API Call Result.'
- Use clear and concise language, phrasing the answer in a way that feels conversational 
  and human-like.
- Ensure the final response directly addresses the user's request without unnecessary 
  detail.
- Summarize or filter the results if needed, prioritizing the most relevant information.
- If any subinstruction does not yield meaningful data, exclude it from the final answer 
  and focus on the most relevant results.
Output Format:
- Return a dictionary with the following keys:
- "thought": Provide a concise summary of how the API Call Result was analyzed, 
  how relevant subinstructions were chosen, and how they were combined to address 
  the User Query.
- "final_answer": A natural language response to the User Query, synthesized from the 
  API Call Result. 
This should sound as if an agent is directly responding to the user.
DO NOT include API jargon or technical terms in the final answer. 
Only present information in user-friendly, natural language.
Focus on delivering the information as if you are the final point of communication with 
the user.
ONLY return the output dictionary. DO NOT include any other words.
\end{verbatim}
\end{tcolorbox}
\caption{\textbf{Final response generation prompt.}}
\label{fig:finalresponse-prompt}
\end{figure*}
\begin{figure*}[t]
\begin{tcolorbox}[colframe=gray!50!black, colback=gray!10!white, title=Success Rate Prompt, width=\textwidth, sharp corners=southwest]
\scriptsize
\begin{verbatim}
Given a user query, a sequence of tool execution details (including successes and 
failures), 
and the final answer, determine whether the answer sufficiently and correctly solves 
the original query, strictly based on the tool execution results.
Evaluation Rules:
1. The final answer must be based on the tool execution results.
- If the answer is generated independently without using the tool results, return 
  "Unsolved".
2. The final answer must address and resolve **all parts** of the user query. 
   Partial answers are not accepted.
- If the answer does not fully respond or give valid answer to every part of the query, 
  return "Unsolved".
3. Only if the answer is fully based on tool results **and** correctly answers all 
   aspects of the query, return "Solved".
   No "Unsure" status is allowed.
Output format:
{
"content": "<Step-by-step reasoning and explanation>",
"answer_status": "Solved" | "Unsolved"
}
\end{verbatim}
\end{tcolorbox}
\caption{\textbf{Prompt used for success rate metric.}}
\label{fig:Success Rate Prompt}
\end{figure*}

\end{document}